\pdfoutput=1
\documentclass[letterpaper]{article} %
\usepackage{aaai25}  %
\usepackage{times}  %
\usepackage{helvet}  %
\usepackage{courier}  %
\usepackage[hyphens]{url}  %
\usepackage{graphicx} %
\usepackage{natbib}  %
\usepackage{caption} %
\usepackage{algorithm}

\usepackage{url}
\usepackage{booktabs}
\usepackage{amsfonts}
\usepackage{enumerate}
\usepackage{mathrsfs}
\usepackage{listings}
\usepackage{mathtools}
\usepackage{todonotes}
\usepackage{adjustbox}
\usepackage{extarrows}
\usepackage{scrextend}
\usepackage{multirow}
\usepackage{makecell}
\usepackage{pifont}
\usepackage{array}
\usepackage{algpseudocode}
\usepackage{cleveref}

\newcommand{\bc}{\mathbf{c}}
\newcommand{\bx}{\mathbf{x}}
\newcommand{\bz}{\mathbf{z}}
\newcommand{\by}{\mathbf{y}}

\newcommand{\bh}{\mathbf{h}}

\newcommand\Exp{\mathbb{E}}
\newcommand\KL{\text{KL}}
\newcommand\Lc{\mathcal{L}}
\newcommand\Uc{\mathcal{U}}

\newcommand{\btheta}{{\boldsymbol{\theta}}}
\newcommand{\bphi}{{\boldsymbol{\phi}}}

\newcommand{\bzero}{\mathbf{0}}

\newcommand{\bI}{\mathbf{I}}

\definecolor{projectlink}{RGB}{200,93,49}
\newcommand{\projecturl}[1]{\leavevmode\pdfstartlink attr{/Border[0 0 0]} user{/Subtype/Link/A<</Type/Action/S/URI/URI(#1)>>}\textcolor{projectlink}{\url{#1}}\pdfendlink}

\usepackage{stackengine}

\let\oldtheta\theta
\renewcommand{\theta}{{\boldsymbol \oldtheta}}

\let\oldpsi\psi
\renewcommand{\psi}{{\boldsymbol \oldpsi}}

\let\oldphi\phi
\renewcommand{\phi}{{\boldsymbol \oldphi}}

\let\oldtau\tau
\renewcommand{\tau}{{\boldsymbol \oldtau}}

\let\oldomega\omega
\renewcommand{\omega}{{\boldsymbol \oldomega}}

\let\oldalpha\alpha
\renewcommand{\alpha}{{\boldsymbol \oldalpha}}

\let\oldepsilon\epsilon
\renewcommand{\epsilon}{{\boldsymbol \oldepsilon}}

\usepackage{newfloat}
\usepackage{listings}
\DeclareCaptionStyle{ruled}{labelfont=normalfont,labelsep=colon,strut=off} %
\floatstyle{ruled}
\newfloat{listing}{tb}{lst}{}
\floatname{listing}{Listing}
\copyrighttext{Code: \projecturl{https://github.com/yingheng-wang/DDVI}}  %
\title{Denoising Diffusion Variational Inference: \\ Diffusion Models as Expressive Variational Posteriors}
\author{
    Yingheng Wang\equalcontrib\textsuperscript{\rm 1},
    Wasu Top Piriyakulkij\equalcontrib\textsuperscript{\rm 1},
    Volodymyr Kuleshov\textsuperscript{\rm 1,2}
}
\affiliations{
    \textsuperscript{\rm 1}Department of Computer Science, Cornell University\\
    \textsuperscript{\rm 2}The Jacobs Technion-Cornell Institute, Cornell Tech\\

    \{wp237, yw2349, kuleshov\}@cornell.edu
}

\begin{document}

\maketitle

\begin{abstract}
We propose denoising diffusion variational inference (DDVI), a black-box variational inference algorithm for latent variable models which relies on diffusion models
as flexible approximate posteriors. 
Specifically, our method introduces
an expressive class of diffusion-based variational posteriors that perform iterative refinement in latent space; we train these posteriors with a novel regularized evidence lower bound (ELBO) on the marginal likelihood inspired by the wake-sleep algorithm. 
Our method is easy to implement (it fits a regularized extension of the ELBO), is compatible with black-box variational inference, and generally outperforms alternative classes of approximate posteriors based on normalizing flows or adversarial networks.
We find that DDVI improves inference and learning in deep latent variable models across common benchmarks as well as on a motivating task in biology---inferring latent ancestry from human genomes---where it outperforms strong baselines on the 1000 Genomes dataset.
Code and blogpost are on the project page:\par\vspace{-\parskip}
{\centering\projecturl{https://yingheng-wang.com/DDVI}\par}
\end{abstract}

\section{Introduction}

We are interested in amortized black-box variational inference 
problems of the form
\begin{align}
    \log\; p_\theta(\bx) \geq &\max_\phi \;\; \Exp_{q_\phi(\bz|\bx)} \left[ \log \; p_\theta(\bx, \bz) - \log\; q_\phi(\bz|\bx) \right] \nonumber \\ 
    &:= \max_\phi \;\; \text{ELBO}(\bx, \theta, \phi), 
\end{align}
in which we approximate the marginal likelihood $\log\; p_\theta(\bx)$ of a latent variable model $p_\theta(\bx, \bz)$ with an evidence lower bound $\text{ELBO}(\bx, \theta, \phi)$ that is a function of an approximate variational posterior $q_\phi (\bz|\bx)$.
We assume that $p_\theta$ factorizes as $p_\theta(\bx|\bz)p_\theta(\bz)$ and admits efficient sampling: examples of such $p_\theta$ include Bayesian networks, topic models \cite{blei2003latent}, variational autoencoders (VAEs), and broad classes of $p_\theta$ defined via modern probabilistic programming frameworks \cite{gordon2014probabilistic}.

Maximizing $\text{ELBO}(\bx, \theta, \phi)$ over $\phi$ yields a variational posterior $q_\phi(\bz|\bx)$ that approximates $p_\theta(\bz|\bx)$ as well as a tight bound on $\log\; p_\theta(\bx)$ that serves as a learning objective for $p_\theta$.
The approximation gap $\log\; p_\theta(\bx) - \max_\phi \text{ELBO}(\bx, \theta, \phi)$ equals precisely $\min_\phi \text{KL}(q_\phi(\bz|\bx) || p_\theta(\bz|\bx))$, which motivates the design of expressive classes of posteriors $q_\phi(\bz|\bx)$ that reduce this gap.
Recent efforts %
leverage modern generative models---including normalizing flows \citep{rezende2015variational,kingma2016improved} and generative adversarial networks \citep{goodfellow2014generative,makhzani2015adversarial}---as expressive model families for $q_\phi$ that tighten the ELBO.

This work seeks to further improve variational inference via expressive posteriors based on diffusion models
\citep{ho2020denoising, song2020score}.
Diffusion methods have become 
the de facto standard for high-quality image synthesis \citep{rombach2022high,gokaslan2024commoncanvas}.
Here, we use diffusion in latent space to parameterize $q_\phi(\bz|\bx)$.
We train this distribution with a denoising diffusion-like objective that does not involve adversarial training \citep{makhzani2015adversarial} or constrained invertible normalizing flow architectures \citep{kingma2016improved}. Samples from $q_\phi(\bz|\bx)$ are obtained via iterative refinement of $\bz$, starting from a Gaussian distribution, and gradually forming one that is multi-modal and complex.

Our work expands upon
existing diffusion-based approximate inference methods \cite{berner2022optimal, zhang2021path, vargas2023denoising, zhang2023diffusion, richter2023improved, sendera2024diffusion, akhound2024iterated} that focus on the task of drawing samples from unnormalized distributions $\tilde p(\bz)$ and estimating the partition function $Z = \int_\bz \tilde p(\bz) d\bz$. 
While these methods are applicable in our setting---we set the unnormalized $\tilde p(\bz)$ to $p_\theta(\bx,\bz)$ such that $Z = p_\theta(\bx)$---they do not make use of characteristics of $p_\theta$ that are common in many types of models (VAEs, Bayes networks, etc.), namely the factorization $p_\theta(\bx|\bz)p_\theta(\bz)$ and efficient sampling.
We find that leveraging these properties yields simpler algorithms that do not rely solely on backpropagating through a sampling process, and that are fast enough to perform learning in addition to inference.

Specifically, we propose denoising diffusion variational inference (DDVI), an approximate inference algorithm defined by a class of approximate posterior distributions based on diffusion and a learning objective inspired by the wake-sleep algorithm \citep{hinton1995wake} that implements regularized variational inference.
We also derive extensions of our method to semi-supervised learning and clustering.

We evaluate DDVI on synthetic benchmarks and on a real problem in biological data analysis---inferring human ancestry from genetic data. Our method outperforms strong baselines on the 1000 Genomes dataset \citep{siva20081000} and learns a low-dimensional latent space that preserves biologically meaningful structure \citep{haghverdi2015diffusion}.

\paragraph{Contributions.} In summary, this work introduces denoising diffusion variational inference, an approximate inference algorithm defined by two components: 
a class of approximate posteriors $q(\bz|\bx)$ parameterized by diffusion,
and a lower bound on the marginal likelihood inspired by wake-sleep.
We complement DDVI with extensions to semi-supervised learning and clustering.
Our method is especially suited for probabilistic programming, representation learning, and dimensionality reduction, where it generally outperforms alternative methods based on normalizing flows and adversarial training.

\begin{figure*}[t]
    \centering
    \includegraphics[width=0.9\textwidth]{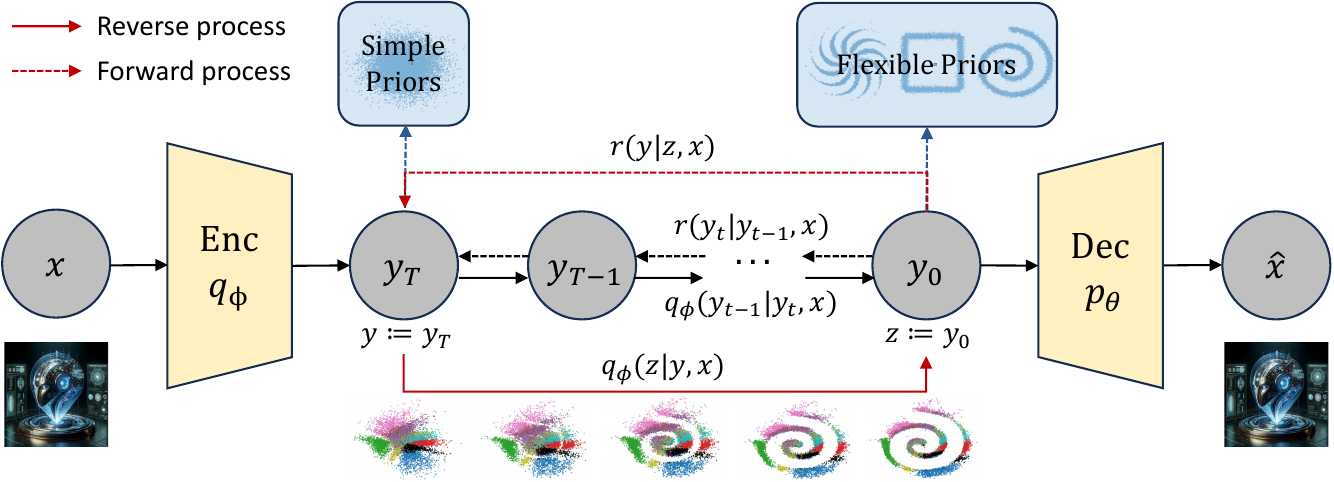}

    \caption{Denoising diffusion variational inference in a VAE. Between the encoder and decoder, we have a diffusion model to map a simple distribution into a complex distribution over latents.}
    \label{fig:diffvae_main}
\end{figure*} 

\section{Background}

\paragraph{Deep Latent Variable Models}

Latent variable models (LVMs)  $p_\theta(\bx, \bz)$ are usually fit by optimizing the evidence lower bound (ELBO) 
\begin{equation}
\log p_\theta(\bx) \geq \mathbb{E}_{q_\phi(\bz|\bx)} [\log p_\theta(\bx|\bz)] - \KL(q_\phi(\bz|\bx)||p_\theta(\bz)), \nonumber
\end{equation}
which serves as a tractable surrogate for the marginal log-likelihood (MLL). The gap between the MLL and the ELBO equals precisely $\KL(q_\phi(\bz|\bx)||p_\theta(\bz|\bx))$---thus, a more expressive $q_\phi(\bz|\bx)$ may better fit the true posterior and induce a tighter ELBO \citep{kingma2013auto}.

Expressive variational posteriors can be formed by choosing more expressive model families---including auxiliary variable methods \citep{maaloe2016auxiliary}, MCMC-based methods \citep{salimans2015mcmc}, normalizing flows \citep{rezende2015variational}---or improved learning objectives---e.g., adversarial or sample-based losses \citep{makhzani2015adversarial,zhao2017infovae,si2022autoregressive,pmlr-v202-si23a}.

The wake-sleep algorithm \citep{hinton1995wake} optimizes an alternative objective
\begin{equation}
\log p_\theta(\bx) - \KL(p_\theta(\bz|\bx)||q_\phi(\bz|\bx)), \nonumber
\end{equation}
in which the KL divergence term of the ELBO, $\log p_\theta(\bx) - \KL(q_\phi(\bz|\bx)||p_\theta(\bz|\bx))$, is reversed. The learning procedure for wake-sleep involves alternating between ``wake'' phases where the generative model is updated on real data and ``sleep'' phases where the recognition model is refined on samples from the generative model.

\paragraph{Denoising Diffusion Models}

A diffusion model is defined via a user-specified noising process $q$ that maps data $\bx_0$ into a sequence of $T$ variables $\by_{1:T} = \by_1, ..., \by_T$ that represent increasing levels of corruption to $\bx_0$. We obtain $\by_{1:T}$ by applying a Markov chain
$q(\by_{1:T} | \bx_0) = \prod_{t=1}^T q(\by_t | \by_{t-1})$, where we define $\by_0 = \bx_0$ for convenience.
When $\bx_0$ is a continuous vector, a standard choice of transition kernel is $q(\by_t \mid \by_{t-1}) = \mathcal{N}(\by_t; \sqrt{\alpha_t} \by_{t-1}, (1-\alpha_t) \mathbf{I})$, which is a Gaussian centered around a copy of $\by_{t-1}$ to which we added noise following a schedule $1 > \alpha_1 > \alpha_2 > \dots > \alpha_T > 0$.

A diffusion model can then be represented as a latent variable distribution $p(\bx_{0}, \by_{1:T})$ %
that factorizes as
$p(\bx_0, \by_{1:T}) = p(\by_{T}) \prod_{t=0}^{T-1} p_\theta(\by_t \mid \by_{t+1})$ (again using $\by_0$ as shorthand for $\bx_0$).
This model seeks to approximate the reverse of the forward diffusion $q$ and map noise $\by_T$ into data $\bx_0$.

The true reverse of the process $q$ cannot be expressed in closed form; as such, we parameterize $p_\theta$ with $\theta$
trained by maximizing the ELBO:
\vspace{-5pt}
\begin{align}
    \log p_\theta(\bx_{0}) \geq &
    \Exp_{q} \Bigr[\log p_\theta(\bx_{0} | \by_{1})  -  \sum_{t=2}^T  \KL(q_t || p_t) \Bigr] \\ \nonumber
    &- \KL(q(\by_T | \bx_{0}) || p(\by_T))
\end{align}
where $q_t, p_t$ denote the distributions $q(\by_{t-1} | \by_{t}, \bx_0)$ and $p_\theta(\by_{t-1} | \by_{t})$, respectively.

\section{Variational Inference with\\Denoising Diffusion Models}
\label{sec:diffvae}

We introduce {\em denoising diffusion variational inference (DDVI)}, which improves variational inference with diffusion-based techniques.

The goal of DDVI is to fit a latent variable model $p_\theta(\bx,\bz)$.
We assume that $p_\theta$ factorizes as $p_\theta(\bx|\bz)p_\theta(\bz)$ and admits efficient sampling: examples of such $p_\theta$ include Bayesian networks and variational autoencoders (VAEs) \cite{kingma2013auto}.

Our approach comprises three components: 
\begin{enumerate}
    \item A \textbf{modeling family} of approximate posteriors $q_\phi(\bz | \bx)$ based on diffusion;
    \item A \textbf{learning objective} formed by a regularized ELBO;
    \item An \textbf{optimization algorithm} inspired by wake-sleep.
\end{enumerate}
The $q_\phi(\bz | \bx)$ iteratively refines latents $\bz$, starting from a Gaussian distribution. 
The learning objective trains $q_\phi(\bz | \bx)$ to reverse a user-specified forward diffusion process.

\subsection{Modeling Family: Diffusion-Based Posteriors}

DDVI performs variational inference using a family of approximate posteriors $q_\phi(\bz|\bx) = \int_\by q_\phi(\bz| \by,\bx) q_\phi(\by|\bx) d\by$, which themselves contain latent variables $\by \in \mathcal{Y}$.
The models $q_\phi(\bz| \by,\bx), q_\phi(\by|\bx)$ must have tractable densities and support gradient-based optimization over $\phi$.

We choose the latent $\by = (\by_1, \by_2, ..., \by_T)$ to be  a vector of $T$ variables that represent progressively simplified versions of $\bz$, with $\by_T$ corresponding to a simple distribution (e.g., a Gaussian).
The model $q_\phi(\by, \bz|\bx) = q_\phi(\by_T|\bx) \, q_\phi(\bz| \by_1,\bx) \prod_{t=1}^{T-1} q_\phi(\by_t|\by_{t+1},\bx)$ transforms $\by_T$ into $\bz$ via iterative refinement.
To sample from $q_\phi$, we first sample $\by_T$---this is an easier task since we can define $\by_T$ to have a simple (e.g., Gaussian) distribution---and then sample from the denoising model $q_\phi(\bz| \by_1,\bx) \prod_{t=1}^{T-1} q_\phi(\by_t|\by_{t+1},\bx)$. 

We define the relationship between $\by$ and $\bz$ via a forward diffusion process $r(\by|\bz,\bx) = r(\by_{1:T}|\bz,\bx) =  r(\by_1|\bz,\bx)\prod_{t=1}^{T-1} r(\by_{t+1}|\by_{t},\bx) $,
which transforms $\bz$---the latent whose intractable posterior we seek to approximate---into $\by_T$, whose posterior is easier to model (possibly conditioned on $\bx$).
Examples of $r$ include Gaussian forward diffusion processes and discrete noising processes \citep{austin2021structured}.
The model $q_\phi$ is trained to approximately reverse this forward diffusion process.

\subsection{Learning Objective: A Markovian ELBO}

The standard approach to fit auxiliary-variable generative models \citep{maaloe2016auxiliary} is to apply the ELBO twice:
\begin{align}
    \log p_\theta(\bx) 
    & \geq \log p_\theta(\bx) - \KL(q_\phi(\bz|\bx)||p_\theta(\bz|\bx)) \label{eqn:elbo2_z1} \\
    & \geq \log p_\theta(\bx) - \KL(q_\phi(\bz|\bx)||p_\theta(\bz|\bx)) \label{eqn:elbo2_y1}  \\ \nonumber
    &~~~~ - \Exp_{q_\phi(\bz|\bx)} [ \KL(q_\phi(\by|\bx,\bz)||r(\by|\bx,\bz)) ] \\
    & = ~\Exp_{q_\phi(\by,\bz|\bx)} [\log p_\theta(\bx|\bz)] \label{eqn:elbo2_y2} \\ \nonumber
    &~~~~ -\KL(q_\phi(\by,\bz|\bx)||r(\by|\bx,\bz)p(\bz))
\end{align}
In Equation (\ref{eqn:elbo2_z1}), we applied the ELBO over $\bz$, and in Equation (\ref{eqn:elbo2_y1}) we applied the ELBO over the latent $\by$ of $q$ (see Appendix~\ref{app:elbo_aux} for the derivation). 
Notice that the gap between the ELBO and $\log p_\theta(\bx)$ is $\KL(q_\phi(\bz|\bx)||p_\theta(\bz|\bx)) + \Exp_{q_\phi(\bz|\bx)} [ \KL(q_\phi(\by|\bx,\bz)||r(\by|\bx,\bz)) ]$. Thus, the bound is tight if $q_\phi(\bz|\bx) = p_\theta(\bz|\bx)$ and $q_\phi(\by|\bx,\bz) = r(\by|\bx,\bz)$.

\subsubsection{Analyzing the ELBO}
Optimizing Equation (\ref{eqn:elbo2_y2}) requires tractably dealing with the prior regularization term $\mathcal{L}_\text{reg}(\bx,\theta,\phi) := -\KL(q_\phi(\by,\bz|\bx)||r(\by|\bx,\bz)p(\bz))$, which we equivalently rewrite as:
\begin{align}
\mathcal{L}_\text{reg} = \Exp_{q_\phi(\by,\bz|\bx)}[\log(r(\by|\bx,\bz)p(\bz))] + H(q). \label{eqn:reg_loss}
\end{align}
We can expand the first term by leveraging the Markov structure of  $r, q$ to rewrite $\mathcal{L}_\text{reg}$ as the likelihood of samples from the reverse diffusion process $q$ under the forward process $r$.
\begin{align}
\mathcal{L}_\text{reg} = \sum_{t=1}^T \Exp_{q}[\log r(\by_t|\by_{t-1},\bx)] + \Exp_{q} [\log p(\bz)] + H(q), \nonumber
\end{align}
where $\by_0 := \bz$. We refer to optimizing the Markovian ELBO as \textbf{unregularized DDVI}, the first instance of our method. 
The noise process $r$ defines prior regularization terms for each $\by_t$. This provides extra supervision for learning $q$ in the form of trajectories from latents $\by_T$ to $\by_1$; this extra supervision helps $q$ fit complex non-Gaussian posteriors.

The term $H(q) = - \sum_{t=1}^{T+1} \mathbb{E}_q [\log q_\phi(\by_{t-1}|\by_t,\bx)]$ denotes the entropy. For example, when each term $q_\phi(\by_{t-1}|\by_t,\bx)$ is Gaussian, it is computed as
\begin{align}
    H(q) = 
    & \sum_{t=1}^{T+1} \mathbb{E}_{q} \left [ \frac{d}{2} \left( 1 + \log(2\pi) \right) + \frac{1}{2} \log \left| \Sigma_\phi(\by_t, \bx) \right| \right ] \nonumber
\end{align}
where $d$ is the dimension of each $\by_t$ and we use the notation $\by_{T+1} = \bx$.
If the variances of the reverse steps are fixed, only the entropy of $q_\phi(\by_T|\bx)$ depends on $\phi$.

\subsection{Refining the Objective: A Regularized ELBO}

Notice that optimizing $\mathcal{L}_\text{reg}$
involves sampling from the approximate reverse process $q_\phi(\by,\bz|\bx)$ to match the forward process $r(\by | \bz, \bx)$: this is the opposite of diffusion training, where we would sample from $r$ to fit $q$. This type of on-policy learning of $q$ has been studied in the context of approximate inference \citep{zhang2021path}; however, it requires backpropagating through $T$ samples, which may hamper training, and it optimizes a mode-seeking divergence that may struggle to fit complex $p(\bz)$. 

\subsubsection{Adding Wake-Sleep Regularization to the ELBO}

We propose to further improve the ELBO via off-policy diffusion-like training. Our new objective is the ELBO in Equation (\ref{eqn:elbo2_y2}) augmented with a regularizer $\Lc_\text{sleep}(\phi)$.
\begin{align} \label{eqn:objective}
\log p_\theta(\bx) \geq &
 \underbrace{\Exp_{q_\phi(\by,\bz|\bx)} [\log p_\theta(\bx|\bz)]}_\text{wake / recons. term $\Lc_\text{rec}(\bx,\theta,\phi)$} \\ \nonumber & \underbrace{-\KL(q_\phi(\by,\bz|\bx)||r(\by|\bx,\bz)p(\bz))}_\text{prior regularization term $\mathcal{L}_\text{reg}(\bx,\theta,\phi)$} \\ \nonumber & \underbrace{-\Exp_{p_\theta(\bx)} [\KL(p_\theta(\bz|\bx)||q_\phi(\bz|\bx))]}_\text{sleep term $\Lc_\text{sleep}(\phi)$}
\end{align}
The optimization of the regularizer $\Lc_\text{sleep}(\phi)$ is similar to the sleep phase of wake-sleep, and closely resembles diffusion model training (see below). 
As in wake-sleep, $\Lc_\text{sleep}(\phi)$ is optimized over $\phi$ only, and the $\bx$ are sampled from $p_\theta$.

\subsubsection{From Wake-Sleep to Diffusion Regularization}

Computing $\Lc_\text{sleep}(\phi)$ still involves intractable distributions $p_\theta(\bz|\bx), q_\phi(\bz|\bx)$.
To optimize $\Lc_\text{sleep}(\phi)$, we introduce another lower bound $\Lc_\text{diff}(\phi)$, which we call the denoising diffusion loss (for reasons that will become apparent shortly):
\begin{align}
\label{eqn:sleep1b}
\Lc_\text{sleep}(\phi) 
& = - \Exp_{p_\theta(\bx)} [\KL(p_\theta(\bz|\bx)||q_\phi(\bz|\bx))] \\ \nonumber
& = \Exp_{p_\theta(\bx,\bz)} [\log {q_\phi(\bz|\bx)}] + \bar H(p_\theta) \\ \nonumber
& \geq \Exp_{p_\theta(\bx,\bz)} \left[ \Exp_{r} [\log \frac{ q_\phi(\by,\bz|\bx)} {r(\by|\bz,\bx) }] \right] + \bar H(p_\theta) \\ \nonumber
& = \Lc_\text{diff}(\phi)
\end{align}
In Equation (\ref{eqn:sleep1b}), we applied the ELBO with $r(\by|\bz,\bx)$ playing the role of the variational posterior over the latent $\by$ in $q_\phi$; $\bar H(p_\theta)$ is the expected conditional entropy of $p_\theta(\bz|\bx)$, a constant that does not depend on $\phi$.

We can further simplify $\Lc_\text{diff}(\phi)$ by leveraging the Markov structure of the forward and reverse processes $r, q$.
Recall that each $\by = (\by_1, \by_2, ..., \by_T)$ can be a vector of $T$ latents, which we also denote as $\by_{1:T}$,
and that
$
r(\by_{1:T}|\bz,\bx) =  \prod_{t=1}^T r(\by_t|\by_{t-1},\bx) 
$,
where $\by_0 = \bz$ %
and also
$
q_\phi(\by, \bz | \bx) = q_\phi(\by_{0:T} | \bx) = q_\phi(\by_T | \bx) \prod_{t=1}^T q_\phi(\by_{t-1}|\by_{t},\bx).
$

We may use the Markov structure in $q, r$ to rewrite $\Lc_\text{diff}(\phi)$ as a sum of $T$ terms, one per Markov step. The derivation is identical to that used to obtain the ELBO of a diffusion model \citep{sohl2015deep}, and yields an expression of the same form:
\begin{align}
\Lc_\text{diff}(\phi) = &~\Exp_{p_\theta(\bx,\bz)} \Exp_{r} \Bigl[\log q_\phi(\bz | \by_{1}, \bx)  -  \sum_{t=2}^T  \KL(r_t || q_t) \\ \nonumber
&- \KL(r(\by_T | \bz, \bx) || q_\phi(\by_T | \bx)) \Bigr] + \bar{H}(p_\theta),
\end{align}
where $r_t,q_t$ denote the distributions $r(\by_{t-1} | \by_{t}, \by_0, \bx)$ and $q_\phi(\by_{t-1} | \by_{t}, \bx)$ (see Appendix~\ref{app:diff_reg} for the derivation).

\subsubsection{Regularized DDVI Objective}

We define the full DDVI objective $\Lc_\text{ddvi}$ to be the sum of the aforementioned terms:
\begin{align}
    \Lc_\text{ddvi}(\bx, \theta, \phi) = \Lc_\text{rec}(\bx, \theta, \phi) + \Lc_\text{reg}(\bx, \theta, \phi) + \Lc_\text{diff}(\phi) \nonumber
\end{align}
Terms $\Lc_\text{reg}$ and $\Lc_\text{diff}$ may be weighted by hyper-parameters $\beta_\text{reg}, \beta_\text{diff} > 0$, as in the $\beta$-VAE framework. The weights used in our experiments are given in Appendix~\ref{app:training_details}.
Note that since $\Lc_\text{diff} \leq \Lc_\text{sleep} \leq 0$, $\Lc_\text{ddvi}(\bx, \theta, \phi)$ (for $\beta_\text{reg} \geq 1$) is a valid lower bound on $\log p_\theta(\bx)$ that is tight when $\beta_\text{reg} = 1$ and $q_\phi(\by, \bz|\bx') = r(\by|\bz,\bx') p_\theta(\bz|\bx')$ for all $\bx'$ ($\Lc_\text{sleep}$ averages over $\bx' \sim p_\theta$).

\subsection{Optimization: Extending Wake-Sleep}\label{sec:latent-wake-sleep}

We may optimize $\Lc_\text{ddvi}(\bx, \theta, \phi)$ using gradient descent by alternating between ELBO optimization and taking sleep steps (see Appendix~\ref{app:pseudocode} for full details):
\begin{enumerate}
    \item Sample $\bx$ from data, take gradient step on $\theta, \phi$ optimizing $\Lc_\text{rec}(\bx, \theta, \phi) + \Lc_\text{reg}(\bx, \theta, \phi)$ (the ``wake" step);
    \item Sample $\bz, \bx$ from $p_\theta$ and take a gradient step on $\phi$ optimizing $\Lc_\text{diff}(\phi)$ (the ``sleep" step).
\end{enumerate}
Again, terms may be weighted by $\beta_\text{reg}, \beta_\text{diff} > 0$.
Note that by maximizing $\Lc_\text{diff}(\phi)$, we fit $q_\phi(\bz|\bx)$ to $p_\theta(\bz|\bx)$ via the forward KL divergence; similarly, by optimizing $\Lc_\text{rec} + \Lc_\text{reg}$ (the ELBO), we fit $q_\phi(\bz|\bx)$ to $p_\theta(\bz|\bx)$ via the reverse KL divergence. Thus, optimizing $\Lc_\text{ddvi}(\bx, \theta, \phi)$ encourages $q_\phi(\bz|\bx)$ to approximate $p_\theta(\bz|\bx)$, and when the two are equal and $q_\phi$ reverses $r$, the bound $\Lc_\text{ddvi}$ is tight.

\paragraph{Simplified Wake-Sleep}
We also consider a lightweight algorithm, in which $r(\by|\bz)$ and $q_\phi(\bz|\by)$ do not depend on $\bx$. This scenario admits the following optimization procedure:
\begin{enumerate}
    \item Sample $\bx$ from data and compute gradient on $\theta, \phi$ optimizing $\Lc_\text{rec}(\bx, \theta, \phi) + \Lc_\text{reg}(\bx, \theta, \phi)$.
    \item Sample $\bz$ from $p(\bz)$ and compute gradient on $\phi$ optimizing the latent sleep term of Appendix~\ref{app:sim}, which replaces $\Lc_\text{diff}(\phi)$; take step on weighted sum of both gradients.
\end{enumerate}
In this case, this term requires only sampling from $p(\bz)$, and the entire loss $\hat\Lc$ can be optimized end-to-end using gradient descent. This algorithm is simpler (there is no separate sleep phase); however, $q_\phi(\bz|\bx)$ may not perfectly approximate $p_\theta(\bz|\bx)$ when $r(\by|\bz)$ and $q_\phi(\bz|\by)$ do not depend on $\bx$, hence $\hat\Lc$ may no longer be a tight bound.

\subsubsection{Practical Considerations}

A common type of noising process compatible with this bound when $\bz$ is continuous is Gaussian diffusion, where we define $r(\by_t | \by_{t-1}) = \mathcal{N}(\by_t; \sqrt{\alpha_t} \by_{t-1}, (1-\alpha_t) \bI)$ for a schedule $(\alpha_t)_{t=1}^T$ with $\bar{\alpha}_t = \prod_{s\le t} \alpha_s$. We then adopt the parameterization $q_\phi(\by_{t-1} | \by_t, \bx) = \mathcal{N}(\by_{t-1}; \mu_\phi (\by_t, \bx, t), \Sigma_\phi (\by_t, \bx, t))$.
It is then common to parameterize $q_\phi$ with a noise prediction network $\epsilon_\phi$ \citep{ho2020denoising}; the sum of KL divergences can be approximated by $\mathbb{E}_{t, \epsilon \sim \mathcal{N}(\bzero, \bI)}||\epsilon - \epsilon_\phi(\sqrt{\bar{\alpha}_t} \by_0 + \sqrt{1-\bar{\alpha}_t} \epsilon, \bx, t)||^2$. In practice, we use linear schedules, deterministic DDIM sampling \citep{song2020denoising} to produce the latents, and a Huber loss in the simplified algorithm (Appendix~\ref{app:training_details}). Other extensions include discrete denoising processes \citep{austin2021structured,sahoo2024simple,schiff2024simple}. In the wake-sleep setting, we know both endpoints $\by_T \sim q(\cdot|\bx)$ and $\by_0 = \bz$ of the diffusion process, opening the possibility for applying optimal transport techniques \citep{cuturi2013sinkhorn,de2021diffusion}.

\section{Extensions}

\subsection{Semi-Supervised Learning}\label{sec:semi}

Following \citet{makhzani2015adversarial}, we extend our algorithm to the semi-supervised learning setting where some data points have labels denoted by $l$. We assume the user provides a model of the form $p_\theta(\bx, \by, \bz, l) = p_\theta(\bx|\bz, l)r(\by|\bz, l)p_\theta(\bz|l)p(l)$; we set the variational distributions to $q_\phi(\bz|\bx, \by, l), q_\phi(\by|\bx), q_\phi(l|\bx)$. In this setting, we consider two cases, depending on whether the label is observed \citep{kingma2014semi}. We extend Equation (\ref{eqn:objective}) to incorporate the label $l$ corresponding to a data point as follows:
\begin{align}
    \Lc_\text{semi} =
    &~\Exp_{q_\phi(\by,\bz|\bx, l)} [\log p_\theta(\bx|\bz, l)] \\ \nonumber 
    &-\KL(q_\phi(\by,\bz|\bx, l)||p_\theta(\by,\bz|l)) \\ \nonumber
    &- \Exp_{p_\theta(\bx|l)} \left [ \KL(p_\theta(\bz|\bx, l)||q_\phi(\bz|\bx, l)) \right ]
\end{align}
When the label $l$ cannot be observed, we treat it as a latent variable and modify the learning objective $\Uc_\text{semi} 
    = \sum_l q_\phi(l|\bx)\Lc_\text{semi}(\bx, l, \btheta, \phi) - \KL(q_\phi(l|\bx)||p(l))$. This yields a lower bound on the (label-conditional) log-likelihood of the dataset:
$
    \tilde{\Lc}_\text{semi} = \sum_{(\bx, l) \in L}\Lc_\text{semi}(\bx, l, \btheta, \bphi) + \sum_{\bx \in U}\Uc_\text{semi}(\bx, \btheta, \bphi),
$
where $L$ and $U$ are the sets of data with and without labels, respectively.

We also want to guarantee that all model parameters can be learned in all cases, including $q_\phi(l|\bx)$, such that this posterior can be applied as a classifier during inference. Thus, we combine this bound with a classification loss to form an extended learning objective:
$
    \tilde{\Lc}_{\text{semi}_\alpha} = \tilde{\Lc}_\text{semi} + \alpha \cdot \Exp_{\tilde{p}(\bx, l)} \left[ \log q_\phi (l|\bx) \right]
$

\subsection{Clustering}\label{sec:clustering}
We also extend our algorithm to clustering via two strategies. In the first approach, we simply formulate a model in which $p_\btheta(\bz)$ is a mixture of desired priors. The means of these priors are characterized by $\btheta$. From these means, cluster membership, denoted as $\bc$, can be deduced. This approach requires no alteration to the learning objective of $q_\phi$ (Appendix~\ref{app:training_details}).

Alternatively, the second method retains the original prior, but introduces an additional latent cluster variable $\bc$, a one-hot vector with $\sum_i c_i = 1$. Thus, the model can be specified as $p_\theta(\bx, \by, \bz, \bc) = p_\theta(\bx|\bz, \bc)r(\by|\bz)p_\theta(\bz)p(\bc)$ with $p(\bc) = \operatorname{Cat}(\boldsymbol\pi)$. Consequently, the variational distributions become $q_\phi(\bz|\by, \bc, \bx), q_\phi(\by, \bc|\bx)$. This yields the objective:
\begin{align}
    \Lc_\text{clus} (\bx) =
    &~\Exp_{q_\phi(\by,\bz, \bc|\bx)} [\log p_\theta(\bx|\bz, \bc)] \\ \nonumber
     &- \KL(q_\phi(\by,\bz,\bc|\bx)||p_\theta(\by,\bz,\bc)) \\ \nonumber &- \Exp_{p_\theta(\bx)} \left [ \KL(p_\theta(\bz | \bx)||q_\phi(\bz|\bx)) \right ]
\end{align}
Expectations over few one-hot clusters are enumerated. Otherwise, we backpropagate through relaxed discrete sampling \citep{jang2016categorical,sahoo2023backpropagation}.
\begin{figure*}[htbp!]
    \centering
    \includegraphics[width=0.95\textwidth]{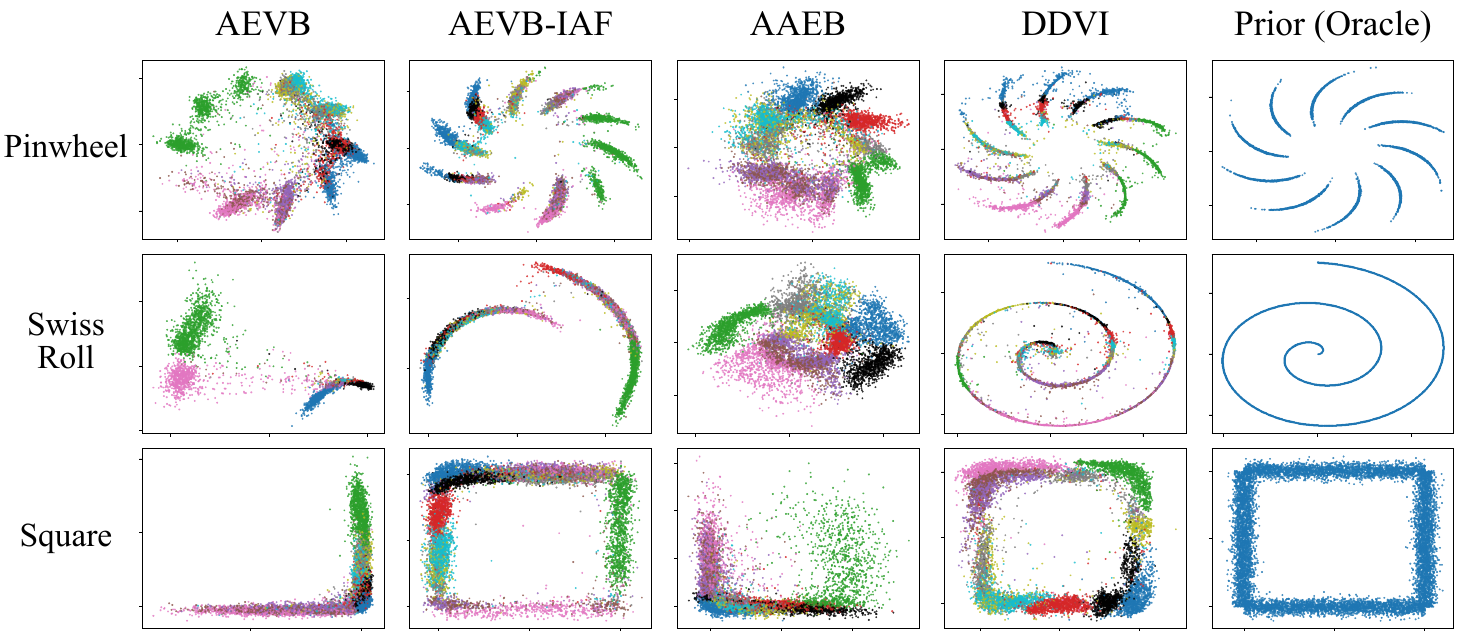}
    
    \caption{Unsupervised visualization on MNIST with three priors (pinwheel, Swiss roll, square). Each color is one class.}
    \label{fig:unsupervised}
\end{figure*} 
\begin{table*}[ht]
\centering
\adjustbox{max width=0.95\textwidth}{
\centering
\setlength\tabcolsep{4pt} %
\begin{tabular}{c@{\hspace{5pt}}c@{\hspace{5pt}}c@{\hspace{5pt}}c@{\hspace{5pt}}c@{\hspace{5pt}}c@{\hspace{5pt}}c@{\hspace{5pt}}c@{\hspace{5pt}}c@{\hspace{5pt}}c}
\toprule
Method & \multicolumn{3}{c}{Pinwheel} & \multicolumn{3}{c}{Swiss Roll} & \multicolumn{3}{c}{Square}\\ 
\cmidrule(lr){2-4} \cmidrule(lr){5-7} \cmidrule(lr){8-10}  
& ELBO & MMD & Latent NLL & ELBO & MMD & Latent NLL & ELBO & MMD & Latent NLL \\ 
\midrule
AEVB & $-169.9 \pm 1.9$ & $0.73 \pm 0.06$ & $1.55 \pm 0.59$ & $-174.6 \pm 2.8$ & $0.74 \pm 0.15$ & $5.63 \pm 1.72$ & $-172.4 \pm 1.2$ & $1.00 \pm 0.27$ & $2.93 \pm 1.11$\\ 
AEVB-IAF & $-159.7 \pm 5.1$ & $0.76 \pm 0.06$ & $1.29 \pm 0.40$ & $-170.7 \pm 17.9$ & $0.76 \pm 0.21$ & $3.70 \pm 1.97$ & $-160.2 \pm 6.7$ & $0.77 \pm 0.16$ & $1.72 \pm 0.10$\\ 
AAEB & $N/A$ & $\mathbf{0.68 \pm 0.02}$ & $1.54 \pm 0.13$ & $N/A$ & $\mathbf{0.54 \pm 0.04}$ & $3.34 \pm 0.24$ & $N/A$ & $0.85 \pm 0.18$ & $2.27 \pm 0.87$\\ 
H-AEVB & $-154.6 \pm 4.0$ & $0.73 \pm 0.07$ & $1.54 \pm 0.11$ & $-158.2 \pm 6.0$ & $0.64 \pm 0.05$ & $2.92 \pm 0.45$ & $-155.8 \pm 7.1$ & $0.75 \pm 0.12$ & $1.67 \pm 0.04$\\ 
PIS$^\dagger$ & $-192.6 \pm 8.2$ & $1.15 \pm 0.04$ & $2.13 \pm 0.65$ & $-216.6 \pm 0.2$ & $1.09 \pm 0.02$ & $6.19 \pm 0.13$ & $-204.6 \pm 0.4$ & $1.49 \pm 0.04$ & $3.68 \pm 0.07$\\ 
DDVI & $\mathbf{-151.5 \pm 2.7}$ & $0.70 \pm 0.08$ & $\mathbf{1.11 \pm 0.10}$ & $\mathbf{-155.9 \pm 1.5}$ & $0.69 \pm 0.05$ & $\mathbf{2.27 \pm 0.16}$ & $\mathbf{-146.6 \pm 2.4}$ & $\mathbf{0.64 \pm 0.03}$ & $\mathbf{1.56 \pm 0.01}$\\ 
\bottomrule
\end{tabular}}
\caption{Unsupervised learning on MNIST: ELBO, MMD between generated and test images, and Latent NLL. The configuration of DDVI was selected on a validation split (Appendix~\ref{app:training_details}). $^\dagger$Our reimplementation (Appendix~\ref{app:model_architecture}).}
\label{tab:unsupervised}
\end{table*}
 
\section{Experiments}\label{sec:exp}

We compare DDVI with Auto-Encoding Variational Bayes (AEVB) \citep{kingma2013auto}, AEVB with inverse autoregressive flow posteriors (AEVB-IAF) \citep{kingma2016improved}, and Adversarial Auto-Encoding Bayes (AAEB) \citep{makhzani2015adversarial} on MNIST \citep{lecun1998mnist} and CIFAR-10 \citep{krizhevsky2009cifar} in unsupervised and semi-supervised learning settings, and also on the 1000 Genomes dataset \citep{siva20081000}. In the unsupervised setting, we also compare with Hierarchical Auto-Encoding Variational Bayes (H-AEVB) \citep{ranganath2016hierarchical,vahdat2020nvae} and Path Integral Sampler (PIS) \citep{zhang2021path}. We discuss computational costs in Appendix~\ref{app:computational_cost}. Priors, architectures, and training details are in Appendices~\ref{app:priors}--\ref{app:training_details}. We report 95\% confidence intervals over 3 seeds (10 for semi-supervised MNIST).

\subsection{Unsupervised Learning}

We start with synthetic experiments that are aimed at benchmarking the expressivity of diffusion-based posteriors and their ability to improve fitting $p$, a distribution with a complex structured prior, like one might find in probabilistic programming, scientific analysis, or other applications.
We fit a model $p_\theta(\bx,\bz)$ on the MNIST and CIFAR-10 datasets with three priors $p(\bz)$: pinwheel, Swiss roll, and square, and report our results in Tables \ref{tab:unsupervised} and \ref{tab:unsupervised_cifar}. 
The model distribution $p_\theta$ is instantiated by a deep Gaussian latent variable model (DGLVM) with multi-layer perceptrons (MLPs) on MNIST and convolutional neural networks (CNNs) on CIFAR-10 (see the details of model architecture in Appendix~\ref{app:model_architecture}).

Our first set of metrics (ELBO and maximum mean discrepancy, MMD) seeks to evaluate whether the learned generative model $p_\theta$ is good. The ELBO is in nats per image (Appendix~\ref{app:training_details}).\footnote{As is common, MNIST uses the Bernoulli log-likelihood of grey-level pixels; $\log p(\bz)$ is a clipped prior KDE (Appendix~\ref{app:priors}); DDVI's $\log q_\phi(\bz|\bx)$ assumes an injective DDIM map; and terms are conditional on dropout masks where dropout is active. The values are thus not guaranteed lower bounds on $\log p_\theta(\bx)$.}
We use MMD to measure sample quality: we generate images with the trained model and calculate MMD between the generated images and test images using a mixture of Gaussian kernels with $\sigma \in \{2, 5, 10, 20, 40, 80\}$. We only report MMD for MNIST, since, with a 2-dimensional latent, CIFAR-10 samples are of very low quality for all methods.

Our last metric seeks to directly evaluate the expressivity of the posterior.
We measure latent negative log-likelihood (Latent NLL) by fitting a kernel density estimator (KDE) on the latents produced by the model with test data as input and computing the negative log-likelihood of prior samples under the fitted KDE.

In Tables \ref{tab:unsupervised} and \ref{tab:unsupervised_cifar}, DDVI has the best ELBO and Latent NLL with all priors on MNIST, and on CIFAR-10 the best Latent NLL with all priors and the best ELBO with square (H-AEVB is higher with the other two, as are AEVB and AEVB-IAF with Swiss roll). AAEB has the best MMD with pinwheel and Swiss roll.
Figure \ref{fig:unsupervised} shows MNIST latents.

\begin{figure*}[t]
    \centering

    \includegraphics[width=0.95\textwidth]{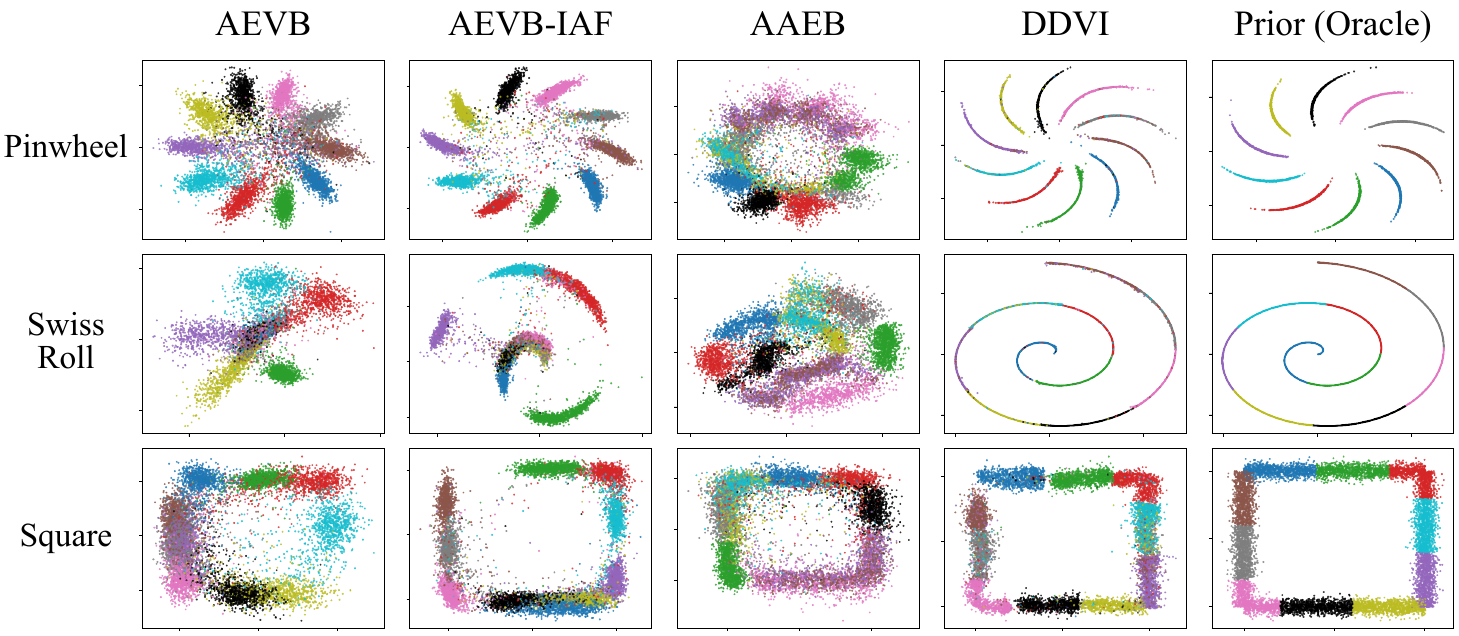}
    \caption{Semi-supervised visualization on MNIST (1,000 labels) with the three priors. Each color is one class.}
    \label{fig:semi-supervised}
\end{figure*}
 
\begin{table*}[ht]
\centering
\adjustbox{max width=\textwidth}{
\centering
\setlength\tabcolsep{4pt} %
\begin{tabular}{c@{\hspace{5pt}}c@{\hspace{5pt}}c@{\hspace{5pt}}c@{\hspace{5pt}}c@{\hspace{5pt}}c@{\hspace{5pt}}c@{\hspace{5pt}}c@{\hspace{5pt}}c@{\hspace{5pt}}c}
\toprule
Method & \multicolumn{3}{c}{Pinwheel} & \multicolumn{3}{c}{Swiss Roll} & \multicolumn{3}{c}{Square}\\ 
\cmidrule(lr){2-4} \cmidrule(lr){5-7} \cmidrule(lr){8-10}  
& ELBO & Acc & Latent NLL & ELBO & Acc & Latent NLL & ELBO & Acc & Latent NLL \\ 
\midrule
AEVB & $-180.7 \pm 0.5$ & $0.85 \pm 0.02$ & $1.59 \pm 0.03$ & $-200.0 \pm 2.0$ & $0.49 \pm 0.02$ & $5.37 \pm 0.22$ & $-191.2 \pm 2.5$ & $0.58 \pm 0.04$ & $2.34 \pm 0.21$\\ 
AEVB-IAF & $-168.0 \pm 0.5$ & $\mathbf{0.90 \pm 0.01}$ & $1.13 \pm 0.03$ & $-175.0 \pm 1.3$ & $0.68 \pm 0.02$ & $4.39 \pm 0.16$ & $-173.6 \pm 0.9$ & $\mathbf{0.82 \pm 0.02}$ & $1.55 \pm 0.03$\\ 
AAEB$^\ddagger$ & $N/A$ & $0.64 \pm 0.02$ & $1.54 \pm 0.03$ & $N/A$ & $0.71 \pm 0.03$ & $3.09 \pm 0.04$ & $N/A$ & $0.59 \pm 0.01$ & $1.80 \pm 0.09$\\ 
DDVI & $\mathbf{-161.9 \pm 0.2}$ & $0.85 \pm 0.01$ & $\mathbf{1.10 \pm 0.01}$ & $\mathbf{-165.8 \pm 0.3}$ & $\mathbf{0.82 \pm 0.01}$ & $\mathbf{2.10 \pm 0.01}$ & $\mathbf{-161.7 \pm 0.3}$ & $\mathbf{0.82 \pm 0.01}$ & $\mathbf{1.51 \pm 0.01}$\\ 
\bottomrule
\end{tabular}}
\caption{Semi-supervised learning on MNIST (1,000 labels): ELBO, 20-NN accuracy (Acc), and Latent NLL. All methods use KL and classification weights 0.01, batch size 128, and 50 epochs ($^\ddagger$except AAEB; Appendix~\ref{app:training_details}).}
\label{tab:semi-supervised}
\end{table*}
 
\subsection{Semi-Supervised Learning}

We also evaluate the performance of our method and the baselines in the semi-supervised setting where some labels are observed (1,000 for MNIST and 10,000 for CIFAR-10) and the partitions of the priors are known.

For this setting, we evaluate ELBO, Latent NLL, and Acc. We choose classification accuracy since classification is a common downstream task for semi-supervised learning. We use the same priors and the AEVB, AEVB-IAF, and AAEB baselines. Details on how we partition each prior into $p_\btheta(\bz|l)$ can be found in Appendix~\ref{app:priors}. The partitions defined for our priors are local parts of the priors. We use one denoiser per partition and, unlike unsupervised learning, the simplified sleep term (Appendices~\ref{app:sim} and~\ref{app:training_details}), since $q_\phi$ already gets extra information from $l$ here. Within each table, all methods except AAEB share the same KL divergence and classification weights (Appendix~\ref{app:training_details}).

On both datasets (Tables \ref{tab:semi-supervised} and \ref{tab:semi-supervised_cifar}), DDVI achieves the best ELBO and Latent NLL with all priors. It also has the best accuracy with all priors on CIFAR-10 and with Swiss roll on MNIST, ties AEVB-IAF with square, and is less accurate with pinwheel.
We also show the visualizations of the latents in Figure \ref{fig:semi-supervised}, where DDVI matches the prior almost perfectly.
\begin{table}[ht]
\centering
\resizebox{\linewidth}{!}{
\begin{tabular}{cccc}
\toprule
Method & Cluster Purity & Cluster Completeness & NMI\\
\midrule
AEVB & $0.28 \pm 0.02$ & $\mathbf{0.78 \pm 0.16}$ & $0.59 \pm 0.08$\\
AEVB-IAF & $0.29 \pm 0.04$ & $0.73 \pm 0.06$ & $0.55 \pm 0.06$\\ 
AAEB & $0.37 \pm 0.06$ & $0.76 \pm 0.11$ & $0.63 \pm 0.02$\\ 
DDVI & $\mathbf{0.45 \pm 0.03}$ & $0.75 \pm 0.05$ & $\mathbf{0.66 \pm 0.04}$\\ 
\bottomrule
\end{tabular}}
\caption{Quantitative genotype clustering results.}
\label{tab:genotype_clustering}
\end{table} 
\subsection{Clustering and~Visualization for Genotype Analysis}

\begin{figure*}[ht]
    \centering

    \includegraphics[width=\textwidth]{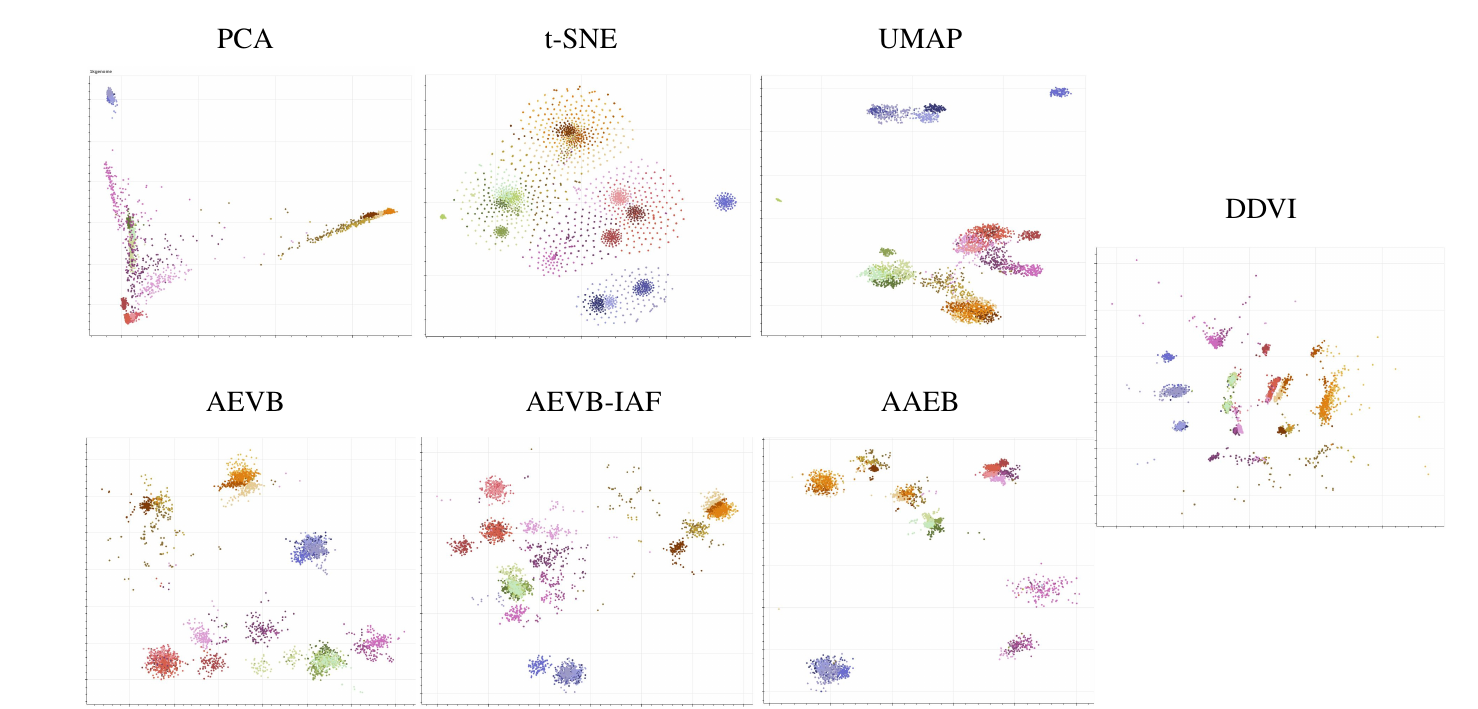}
    \caption{Visualization of genotype clusters. Each color indicates one population.}
    \label{fig:genotype}
\end{figure*}
In this section, we report results on a real-world task in genome analysis. Visualizing genotype data reveals patterns in the latent ancestry of individuals.
We instantiate DDVI with a deep Gaussian latent variable model (DGLVM) and compare it against three strong baselines using the 1000 Genomes dataset.
We also report visualizations from three dimensionality reduction algorithms: PCA \citep{wold1987principal}, t-SNE \citep{van2008visualizing}, and UMAP \citep{mcinnes2018umap}.
For each method, we obtain 20 clusters by fitting a Gaussian mixture model to its latents (Appendix~\ref{app:genotype}). We report quantitative results in terms of cluster purity, cluster completeness, and normalized mutual information (NMI). There is an inherent trade-off between cluster purity and completeness. The overall clustering performance can be captured with NMI.

In Table \ref{tab:genotype_clustering}, we see that DDVI attains the best performance on cluster purity and NMI. For cluster completeness, AEVB and AAEB have better means but much larger confidence intervals.
Figure \ref{fig:genotype} visualizes the latents of all methods (legend in Figure \ref{fig:legend} in the Appendix).

\section{Discussion}

\paragraph{Diffusion vs.~Normalizing Flows}

Our approach is most similar to flow-based approximators \citep{rezende2015variational,kingma2016improved}; in fact, when $T \to \infty$, our diffusion-based posterior effectively becomes a continuous-time normalizing flow \citep{song2020denoising}. However, classical flow-based methods require invertible architectures for each flow layer: this constrains their expressivity and requires backpropagating through a potentially very deep network.
Our approach, on the other hand, trains a model (a continuous-time flow when $T \to \infty$) via a denoising objective (similar to score matching) that does not require invertible architectures and effectively admits an infinite number of layers (with weight sharing). This model is trained not only by backpropagating through the ELBO, but also via an auxiliary diffusion loss term (effectively, a score matching objective). 

Despite training with a modified loss, we observe in Section \ref{sec:exp} that a diffusion model with an expressive denoising architecture yields a higher ELBO than a flow-based posterior (AEVB-IAF) in 11 of the 12 settings. Also, our modified loss based on the forward KL divergence reduces mode collapse (i.e., all modes of the prior are covered well, as reflected by the Latent NLL).

\paragraph{Diffusion vs.~Other Generative Models}

Variational posteriors based on GANs \citep{makhzani2015adversarial} also admit expressive architectures and require only sample-based access to the prior $p(\bz)$. Our diffusion-based approach admits a more stable loss, and is potentially more expressive, as it effectively supports an infinite number of layers (with shared parameters when $T\to \infty$). Unlike GANs, our models also admit explicit likelihoods and allow us to compute the ELBO for model evaluation.
Our approach is similar to variational MCMC \citep{salimans2015mcmc}; however, we train with a better objective augmented with a diffusion loss, and we adopt improved architectures with shared weights across layers.

\paragraph{Diffusion for Approximate Inference}
As noted in the introduction, existing diffusion-based approximate inference methods are applicable in our setting (with $\tilde p(\bz) = p_\theta(\bx,\bz)$ and $Z = p_\theta(\bx)$); moreover, they tackle a harder sampling problem (drawing samples from energy-based models) in more general classes of models (arbitrary unnormalized distributions).
In contrast, we focus on restricted but still important classes of models (VAEs, Bayes networks, etc.), and we solve a broader set of tasks (e.g., maximum-likelihood learning) by using properties of $p_\theta$ (the factorization $p_\theta(\bx|\bz)p_\theta(\bz)$ and efficient sampling from $p_\theta$). 

Our algorithms are also simpler. For example, diffusion sampling methods rely solely on backpropagating through a sampling process to minimize the reverse $\text{KL} (q_\phi || p_\theta)$, which poses challenges with optimization and credit assignment. Some methods based on Schr\"odinger bridges require an iterative optimization process generalizing the Sinkhorn algorithm or computationally expensive on-policy or off-policy \cite{malkin2022trajectory} trajectory-based optimization. In contrast, DDVI complements the ELBO with the forward $\text{KL} (p_\theta || q_\phi)$, optimized with simple gradient-based updates that directly emulate diffusion-based training.

\section{Related Work}

\paragraph{Latent Diffusion}
Latent diffusion models \citep{vahdat2021score,wehenkel2021diffusion,rombach2022high} perform diffusion in the latent space of a VAE.
Their goal is high sample quality, and they introduce into $p$ hierarchical latents with simple Gaussian priors. 

{Our goal is different}: we seek to fit a $p$ with structured latents (e.g., in probabilistic programming or in science applications, users introduce prior knowledge via hand-crafted $p$), and we improve variational inference in this structured model by introducing auxiliary latents into $q$.
Recent work \citep{preechakul2022diffusion, zhang2022unsupervised, wang2023infodiffusion} has also melded auto-encoders with diffusion models, focusing on semantically meaningful low-dimensional latents in a diffuser $p$. 
\citet{cohen2022diffusion} craft a diffusion bridge linking a continuous coded vector to a non-informative prior distribution.
\paragraph{Diffusion for Approximate Inference}

Diffusion sampling \cite{berner2022optimal, zhang2021path, vargas2023denoising, zhang2023diffusion, richter2023improved, sendera2024diffusion, akhound2024iterated} mainly focuses on 
the task of drawing samples from unnormalized distributions and estimating the partition function. 
These works draw connections between diffusion (learning the denoising process) and stochastic control (learning the Föllmer drift).
Some other works \cite{zhang2023diffusion, sendera2024diffusion} use continuous generative flow networks (GFlowNets)---deep reinforcement learning algorithms adapted to variational inference that offer stable off-policy training and thus flexible exploration. 
While this work is applicable to our setting, it does not rely on the structure of $p(\bx,\bz)$ available to us, namely tractable sampling in $p$.

\paragraph{Dimensionality Reduction}
Latent variable models in general are an attractive alternative to visualization methods like PCA, Sparse PCA, NMF, UMAP, and t-SNE \citep{wold1987principal,kuleshov2013fast,NIPS2000_f9d11525,mcinnes2018umap,van2008visualizing}. Domain-specific knowledge can be injected through the prior, and deep neural networks can be utilized to achieve a more expressive mapping from the data space to the latent space. Nevertheless, downsides of LVMs are that they are more computationally expensive and require careful hyperparameter tuning. %

\section{Conclusion}

While this paper focuses on applications of DDVI to dimensionality reduction and visualization, there exist other tasks for the algorithm, e.g., density estimation or sample generation. 
Accurate variational inference has the potential to improve downstream applications of generative modeling, e.g., decision making \citep{DBLP:conf/icml/NguyenG22,deshpande2023calibrated}, meta-learning \citep{rastogi2023semi}, or causal effect estimation \citep{deshpande2022deep}.

Since our learning objective differs from the ELBO (it adds a regularizer), we anticipate gains on models whose training benefits from regularization, but perhaps not on all models. Also, attaining competitive likelihood estimation requires architecture improvements that are orthogonal to this paper. However, our ability to generate diverse samples and achieve class separation in latent space hints at the method's potential on these tasks.

\newpage

\onecolumn

\title{Appendix}
\maketitle

\appendix
\section{Pseudocode}\label{app:pseudocode}
Here we provide pseudocode to illustrate the training process of DDVI. 
\begin{algorithm*}[h]
\caption{DDVI Pseudocode}
\begin{algorithmic}[1]
\State (Optional) Pre-train $p_\theta(\bx|\bz)$ and $q_\phi(\by|\bx)$ with DDVI but with an unconditional diffusion model $q_\phi(\bz|\by)$
\For{epoch $= 1, \ldots, n$}
    \For{$\bx_1, \ldots, \bx_k \sim p_\mathcal{D}(\bx)$}
        \State $\by_i \sim q_\phi(\by|\bx_i)$ and $\bz_i \sim q_\phi(\bz|\by_i, \bx_i)$ for $i = 1, \ldots, k$
        \State Optimize $\theta, \phi$ with respect to a Monte Carlo estimate of $\mathbb{E}_{q_\phi(\by,\bz|\bx)} \left[ \log p_\theta(\bx|\bz) \right] - \mathrm{KL}\left(q_\phi(\by,\bz|\bx) \| p_\theta(\by,\bz) \right)$ for each $\bx_i$ \Comment{Standard ELBO training part}
        \For{iteration $= 1, \ldots, m$} \Comment{Do sleep for $m$ iterations}
            \State $\bz_{1}, \ldots, \bz_{k} \sim p(\bz)$ \Comment{Batch-sample latents from prior}
            \State $\hat{\bx}_i \sim p_\theta(\bx|\bz_i)$ for $i = 1, \ldots, k$ \Comment{Construct fantasy inputs}
            \State $\by_i \sim r(\by|\bz_i)$ for $i = 1, \ldots, k$ \Comment{Noise the latents with the forward process}
            \State Optimize $\phi$ using the standard diffusion noise prediction loss on $q_\phi(\bz|\by_i, \hat{\bx}_i)$
        \EndFor
    \EndFor
\EndFor
\end{algorithmic}
\label{alg:ddvi}
\end{algorithm*}
 
\section{Simplifying Wake-Sleep}\label{app:sim}
In wake-sleep, sampling $\bx$ from $p_\theta$ to obtain gradients for the sleep term introduces computational overhead. To address this issue, we propose {\em wake-sleep in latent space}, an algorithm that optimizes an approximation $\hat \Lc(\bx,\theta,\phi)$ of $\Lc_\text{ddvi}$:
\begin{align}
    \hat \Lc(\bx, \theta, \phi)
     & =
    \underbrace{\Exp_{q_\phi(\by,\bz|\bx)} [\log p_\theta(\bx|\bz)]}_\text{wake / reconstr. term $\Lc_\text{rec}(\bx,\theta,\phi)$} \underbrace{-\KL(q_\phi(\by,\bz|\bx)||p_\theta(\by,\bz))}_\text{prior regularization term $\mathcal{L}_\text{reg}(\bx,\theta,\phi)$}
    \underbrace{-\KL(p_\theta(\bz)||q_\phi(\bz|\bx))}_\text{latent sleep term $\Lc_\text{sleep}(\bx,\phi)$}.
    \label{eqn:objective2}
\end{align}
We have replaced $\Lc_\text{sleep}(\phi)$ with a latent sleep term $\Lc_\text{sleep}(\bx,\phi)$, in which $\bx$ is given, and we only seek to fit the true reverse noising process $r(\bz|\by)$ independently of $\bx$. We can similarly show that
\begin{align}
\Lc_\text{sleep}(\bx, \phi) 
& = \Exp_{p_\theta(\bz)} [\log {q_\phi(\bz|\bx)}] + \bar H(p_\theta)
 \geq \Exp_{p_\theta(\bz) r(\by|\bz)} [\log ({q_\phi(\by,\bz|\bx)}/r(\by|\bz))] + \bar H(p_\theta) \label{eqn:sleep2} \\
 & = -\Exp_{p_\theta(\bz)} [\KL(r(\by|\bz)||q_\phi(\by|\bz,\bx))] - \KL(p_\theta(\bz)||q_\phi(\bz|\bx)), \label{eqn:sleep2_kl}
\end{align}
where $\bar H(p_\theta)$, the entropy of $p_\theta(\bz)$, is constant in $\phi$.
Thus, we minimize the forward KL divergence 
by sampling $\bz$, and applying the noising process to get $\by$; the $q_\phi$ is fit to denoise $\bz$ from $\by$ as in Equation~(\ref{eqn:sleep1b}).

We optimize our bound on $\hat \Lc(\bx,\theta,\phi)$ end-to-end using  minibatch gradient descent over $\theta, \phi$. 
While the wake term is a reconstruction loss as in wake-sleep, the sleep term generates latent samples $\bz, \by$ from $r(\by|\bz)p_\theta(\bz)$ (by analogy with $p_\theta(\bx|\bz)p_\theta(\bz)$ in normal wake-sleep); the denoiser $q_\phi$ is trained to recover $\bz$ from $\by$.
Thus, we perform wake-sleep {\em in latent space}, 
which obviates the need for alternating wake and sleep phases, and allows efficient end-to-end training. A limitation of this approximation is that the sleep term does not fit $q_\phi$ to the true $p_\theta(\bz|\bx)$, and as a consequence $\hat \Lc$ is not a tight lower bound on $\log p_\theta(\bx)$. We may think of $\Lc_\text{sleep}(\bx, \phi)$ as a regularizer to the ELBO.
Table~\ref{tab:unsupervised_w_failed} illustrates the importance of the sleep term: removing the sleep step from the second training stage of DDVI (all other settings unchanged) yields a substantially worse Latent NLL.

\section{Comparison of Methods}
We provide a comprehensive comparison of different methods in Table \ref{tab:method_comparsion}.
Latent diffusion models \citep{vahdat2021score,wehenkel2021diffusion,rombach2022high} perform diffusion in the latent space of a VAE to improve the efficiency of image generation.
Their goal is high sample quality, and they introduce into $p$ hierarchical latents with simple Gaussian priors. \textbf{Our goal is different}: we seek a method to fit a $p$ with structured latents (e.g., in probabilistic programming or in science applications, users introduce prior knowledge via hand-crafted $p$), and we improve variational inference in this structured model by introducing auxiliary latents into $q$.

Recent work \citep{preechakul2022diffusion, zhang2022unsupervised, wang2023infodiffusion} has also melded auto-encoders with diffusion models, focusing on semantically meaningful low-dimensional latents in a diffuser $p$. 
\citet{cohen2022diffusion} craft a diffusion bridge linking a continuous coded vector to a non-informative prior distribution.
\begin{table*}[htbp!]
\adjustbox{max width=\textwidth}{
\centering
\renewcommand{\arraystretch}{1.5}
\setlength\tabcolsep{4pt} %
    \definecolor{OliveGreen}{HTML}{808000}    %
    \definecolor{bloodred}{HTML}{B00000}      %
    \definecolor{cautionyellow}{HTML}{EED202} %

    \newcommand*\colourcheck[1]{%
      \expandafter\newcommand\csname #1check\endcsname{%
        \textcolor{#1}{\ding{51}}%
      }%
    }
    
    \newcommand*\colourxmark[1]{%
      \expandafter\newcommand\csname #1xmark\endcsname{%
        \textcolor{#1}{\ding{55}}%
      }%
    }
    
    \newcommand*\colourcheckodd[1]{%
      \expandafter\newcommand\csname #1checkodd\endcsname{%
        \textcolor{#1}{\ding{53}}%
      }%
    }

    \colourcheck{OliveGreen}       %
    \colourcheck{cautionyellow}    %
    \colourcheckodd{cautionyellow}%
    \colourxmark{bloodred}         %
    
    \newcommand{\ourxmark}{\bloodredxmark}%
    \newcommand{\ourcheckmark}{\OliveGreencheck}%

    \newcolumntype{C}[1]{>{\centering\arraybackslash}p{#1}}
\begin{tabular}{C{2cm} C{2.3cm} C{2.6cm} C{2cm} C{1.5cm} C{3cm} C{3.5cm}}
\toprule
Model & Training Objective & Approximating Family & Sample-based Prior & Auxiliary Variable & Tasks & Simplified Graphical Illustration \\
\midrule
AEVB & ELBO & Diagonal Gaussian & \ourxmark & \ourxmark & Density estimation & $\mathbf{x} \rightarrow \mathbf{z} \rightarrow \mathbf{x}$ \\
AEVB-IAF & ELBO & Normalizing flow & \ourxmark & \ourcheckmark & Density estimation / Visualization & $\mathbf{x} \rightarrow \mathbf{z}_0 \rightarrow \mathbf{z}_T \rightarrow \mathbf{x}$ \\
AAEB & Adversarial training & Adversarial generator & \ourcheckmark & \ourxmark & Visualization & $\mathbf{x} \rightarrow \mathbf{z} \rightarrow \mathbf{x}$ \\
H-AEVB & ELBO & Factorial Normal / Normalizing flow & \ourxmark & \ourcheckmark & Density estimation / High-quality sample generation & $\mathbf{x} \rightarrow \mathbf{z}_0 \rightarrow \mathbf{z}_T \rightarrow \mathbf{z}_0 \rightarrow \mathbf{x}$\\
ADGM & ELBO & Non-Gaussian & \ourxmark & \ourcheckmark & Density estimation & $\mathbf{x} \rightarrow \mathbf{a} \rightarrow \mathbf{z} \rightarrow \mathbf{x}$ \\
LDM & ELBO & Diagonal Gaussian & \ourxmark & \ourcheckmark & High-quality sample generation & $\mathbf{x} \rightarrow \mathbf{z}_0 \rightarrow \mathbf{z}_T \rightarrow \mathbf{z}_0 \rightarrow \mathbf{x}$ \\
LSGM & ELBO \& score matching & Diagonal Gaussian & \ourxmark & \ourcheckmark & High-quality sample generation & $\mathbf{x} \rightarrow \mathbf{z}_0 \rightarrow \mathbf{z}_T \rightarrow \mathbf{z}_0 \rightarrow \mathbf{x}$ \\
DDVI & Wake ELBO + sleep term & Denoising diffusion & \ourcheckmark & \ourcheckmark & Density estimation / Visualization & $\mathbf{x} \rightarrow \mathbf{z}_T (\mathbf{y}) \rightarrow \mathbf{z}_0 (\mathbf{z}) \rightarrow \mathbf{x}$ \\
\bottomrule
\end{tabular}}
\caption{Comparison of DDVI to other relevant methods. $\mathbf{x}$ represents the original data input to the model. $\mathbf{z}$ denotes the latent (hidden) representation of the input data. $\mathbf{a}$ represents an auxiliary variable introduced in some models (like ADGM) to capture additional aspects of the data distribution or to assist in the model's learning process.}
\label{tab:method_comparsion}
\end{table*} 
\section{Computational Cost Analysis}
\label{app:computational_cost}

\begin{table*}[ht]
\adjustbox{max width=\textwidth}{
\centering
\setlength\tabcolsep{4pt} %
\begin{tabular}{ccccccc}
\toprule
Method & \multicolumn{6}{c}{NMI values at different wall-clock training times}\\
\cmidrule{2-7}
& NMI @ 10 min & NMI @ 20 min & NMI @ 30 min & NMI @ 40 min & NMI @ 50 min & NMI @ 60 min \\
\midrule
AEVB             & $0.52$       & $0.52$       & $0.52$       & $0.52$       & $0.52$       & $0.52$       \\
AEVB-IAF         & $0.54$       & $0.52$       & $0.52$       & $0.52$       & $0.52$       & $0.52$       \\
AAEB             & $0.61$       & $0.57$       & $0.57$       & $0.57$       & $0.57$       & $0.57$       \\
DDVI ($T=5$) & \textit{warm-up} & $0.63$ & $0.63$ & $0.66$ & $0.66$ & $0.66$\\ 
DDVI ($T=10$) & \textit{warm-up} & $\mathbf{0.64}$ & $\mathbf{0.68}$ & $\mathbf{0.70}$ & $\mathbf{0.70}$ & $\mathbf{0.70}$\\ 
DDVI ($T=20$) & \textit{warm-up} & $0.50$ & $0.51$ & $0.56$ & $0.64$ & $0.68$\\ 
DDVI ($T=50$) & \textit{warm-up} & $0.52$ & $0.54$ & $0.51$ & $0.59$ & $0.59$\\ 
\bottomrule
\end{tabular}}
\caption{Computational cost trade-off on 1000 Genomes: NMI vs.\ wall-clock training time.}
\label{tab:computational_cost}
\end{table*} 
We compare the computational cost of the baselines and of DDVI with different numbers of diffusion steps $T$ on the genotype clustering and visualization experiments. Table \ref{tab:computational_cost} shows that, after a short warm-up, DDVI with $T \in \{5, 10\}$ outperforms the baselines at every later wall-clock budget, and that DDVI continues to improve after the baselines have plateaued.

\section{Connections to Diffusion Samplers}
Diffusion sampling \cite{berner2022optimal, zhang2021path, vargas2023denoising, zhang2023diffusion, richter2023improved, sendera2024diffusion, akhound2024iterated} mainly focuses on 
the task of drawing samples from unnormalized distributions and estimating the partition function. 
These works draw connections between diffusion (learning the denoising process) and stochastic control (learning the Föllmer drift), leading to several approaches, e.g., path integral sampler (PIS) \cite{zhang2021path}, denoising diffusion sampler (DDS) \cite{vargas2023denoising}, and time-reversed diffusion sampler (DIS) \cite{berner2022optimal}, which have been unified by \citet{richter2023improved}. Some other works \cite{zhang2023diffusion, sendera2024diffusion} use continuous generative flow networks (GFlowNets)---deep reinforcement learning algorithms adapted to variational inference that offer stable off-policy training and thus flexible exploration. \citet{sendera2024diffusion} benchmarked these previous diffusion-structured amortized inference methods and studied how to improve credit assignment in diffusion samplers, which refers to the propagation of learning signals from the target density to the parameters of earlier sampling steps.  Overall, there are indeed some strong connections between these works and ours:
\begin{itemize}
    \item They also focus on variational methods that directly fit a parametric family of tractable distributions (given by controlled SDEs) to the target density.
    \item They cast the density estimation/sampling problem into an optimization problem over a control objective, which learns control drifts (and diffusion) parameterized by neural networks.
\end{itemize}
There are also clear differences between them:

\begin{itemize}
    \item The diffusion-structured samplers only focus on density estimation/sampling but ignore the problem of learning a generative model, which is one of the main focuses of our work. We aim to perform more accurate variational inference using an auxiliary variable model augmented by diffusion models to improve generative modeling. In our setting, $p_\theta(z|x)$ is a moving target density, as we jointly learn $\theta$ with $\phi$, as opposed to a static target density that diffusion-structured samplers are designed to solve.
    \item To tackle the challenge of credit assignment (propagating weak learning signals through the sampling trajectory), the techniques proposed in diffusion-structured samplers are mostly based on partial trajectory information, which has higher training costs than on-policy \cite{zhang2021path} or off-policy \cite{malkin2022trajectory} trajectory-based optimization. Instead, we introduce a wake-sleep optimization algorithm and its simplified version to alleviate the weak learning signal issue and optimize the evidence lower bound in a better way.
    \item In Equation~(\ref{eqn:sleep1b}), we are minimizing the forward KL divergence $\KL (p_\theta || q_\phi)$, whereas diffusion samplers are minimizing the reverse $\KL (q_\phi || p_\theta)$.
\end{itemize}

We also summarize the connections and differences in Table~\ref{tab:comparison}.

\begin{table}[ht]
\centering
\begin{tabular}{@{}p{2cm}p{4cm}p{4cm}p{4cm}@{}}
\toprule
                        & \textbf{Diffusion Samplers} & \textbf{GFlowNet-based} & \textbf{DDVI (ours)} \\ 
                        & \cite{berner2022optimal, zhang2021path, vargas2023denoising, richter2023improved, akhound2024iterated}              & \cite{zhang2023diffusion, sendera2024diffusion, malkin2022trajectory}               &             \\ \midrule
\textbf{Tasks}                   & Sampling, density estimation  & Sampling, density estimation  & Learning, Sampling, Dimensionality reduction \\
\textbf{Model Family for $p$}    & Any energy-based              & Any energy-based              & Latent with tractable $p(x|z), p(z)$ \\
\textbf{Model Family for $q$}    & Markov chain                  & Markov chain                  & Markov chain \\
\textbf{Objective}  & $\KL(q||p)$ with regularizer & Trajectory balance objective & $\KL(p||q)$ with ELBO \\
\textbf{Algorithm}  & Gradient descent (with reference process), importance sampling & RL-motivated off-policy optimization (replay buffers, Thompson sampling, etc.) & Gradient descent with wake-sleep \\
\textbf{Compatible Models} & Anything energy-based  & Anything energy-based  & LDA, deep latent-variable models \\
\textbf{Applications}            & Sampling from physics-based models, model selection based on NLL & Sampling from physics-based models, model selection based on NLL & Probabilistic programming, visualization \\ \bottomrule
\end{tabular}
\caption{Comparison of Diffusion-structured Samplers, GFlowNet-based Approaches, and DDVI}
\label{tab:comparison}
\end{table}

\section{Priors}
\label{app:priors}

Below we describe the sampling process for each prior. 

\textbf{Pinwheel.} This distribution was used by \citet{johnson2016svae}. We define the number of clusters to be 10. For semi-supervised learning experiments, this prior is partitioned into 10 partitions, each partition being a cluster.

\textbf{Swiss Roll.} This distribution was used by \citet{marsland2014swissroll}. For semi-supervised learning experiments, this prior is partitioned into 10 partitions. The samples from the prior can actually be characterized by a single scalar $u\in[0,1]$, with radius $3\sqrt{u}$ and angle $4.5\pi\sqrt{u}$. The partitioning is done by creating 10 equal-length intervals of $u$.

\textbf{Square.} This distribution has the shape of a square going from $-1$ to $1$ in both axes. Each position on the square can be characterized by a single scalar representing the distance along the square from the top-left corner. Sampling is done by sampling the position uniformly and turning the 1D position into a 2D latent. We add Gaussian noise with $\sigma = 0.06$ to the prior. For semi-supervised learning experiments, this prior is partitioned into 10 partitions. The partitioning is done by creating 10 equal-length intervals in the 1D position space.

AEVB, AEVB-IAF, H-AEVB, and PIS require that we can evaluate the prior density, and so do the term $\Exp_{q}[\log p(\bz)]$ in the wake step of DDVI and the ELBO evaluation (Appendix~\ref{app:training_details}). To do this, for all priors, we fit kernel density estimators with Gaussian kernels of bandwidths 0.005, 0.008, 0.01, 0.03, and 0.05 to $10^4$ prior samples and average their densities; each log-density is clipped to $[-1000, 1000]$. The partition densities $p(\bz|l)$ use 2,000 samples per partition, a single bandwidth of 0.05, and clipping to $[-300, 300]$. 

\section{Model Architecture}
\label{app:model_architecture}

All methods use the same architecture for encoder $q_\phi(\bz|\bx)$ and decoder $p_\theta(\bx|\bz)$, excluding the extra parts specific to each method which we describe below, for the same dataset. For MNIST, the encoder and decoder are multi-layer perceptrons with two hidden layers, each with 1000 hidden units. For CIFAR-10, the encoder is a  4-layer convolutional neural network with (16, 32, 64, 128) channels with a linear layer on top, and the decoder is a 4-layer transposed convolutional neural network with (64, 32, 16, 3) channels, preceded by a linear layer that maps the 2-dimensional latent to 64 features and followed by a tanh, so that its outputs lie in $[-1, 1]$, the range of the normalized images (an earlier version of this paper used a sigmoid for all methods).

\textbf{AEVB-IAF} employs 4 IAF transformations on top of the encoder, each implemented with a 4-layer MADE. The number of hidden units in MADE is 128. The ordering is reversed between every other IAF transformation. 

\textbf{H-AEVB} has a two-level posterior. The first level $q_\phi(\bz_1|\bx)$ is as in AEVB-IAF (4 IAF transformations, with 64 hidden units in MADE). The second level $q_\phi(\bz|\bx,\bz_1)$ is a Gaussian whose parameters are computed from features of $\bx$ (an MLP with the same architecture as the encoder for MNIST; the convolutional encoder for CIFAR-10) and from $\bz_1$ by an MLP with two hidden layers of 24 units, followed by another 4 IAF transformations. The decoder takes $\bz$; $\bz_1$ is an auxiliary variable with the reverse model $\mathcal{N}(\bzero,\bI)$.

\textbf{AAEB} has a discriminator, used in adversarial training, which is a multi-layer perceptron with two hidden layers, each with 1000 hidden units.

\textbf{PIS} is our reimplementation of the path integral sampler \citep{zhang2021path} as an amortized variational posterior, since no implementation for this setting is publicly available. The encoder outputs a 10-dimensional context vector, and the drift network has the same architecture as the DDVI denoiser below (a 5-layer time-conditioned multi-layer perceptron with 128 hidden units), taking $\bz$ and the context vector as input.

\textbf{DDVI} has a diffusion model on top of the encoder. The time-conditioned reverse diffusion distribution is implemented with a 5-layer time-conditioned multi-layer perceptron with 128 hidden units per layer. Each time-conditioned linear layer learns an additional embedding for each timestep and adds it to its output. This denoiser does not depend on $\bx$; it is used in the first training stage of the unsupervised experiments and in the semi-supervised and clustering experiments (Appendix~\ref{app:training_details}). In the second training stage of the unsupervised experiments, the reverse process depends on $\bx$: the noise predictor $\epsilon_\phi(\by_t, t, \bh(\bx))$ has 4 residual blocks of width 256 (with LayerNorm and SiLU activation), each modulated (FiLM) by an embedding of $t$ and of a 64-dimensional context $\bh(\bx)$, which is an additional output head of the encoder. The latent is standardized with the mean and standard deviation of the prior, and the predicted $\bz$ (in standardized units) is passed through a smooth clip, the identity on $[-4, 4]$ followed by a tanh saturation to $\pm 5$, which, unlike a hard clamp, has a nonzero derivative everywhere. In this stage, dropout (rate 0.2) in the MLP layers is used during training only.

\section{Training Details} 
\label{app:training_details}

For DDVI, we alternate between the ELBO phase (optimizing $\theta$ and $\phi$ with respect to the ELBO, i.e., the reconstruction term and the prior regularization term) and the sleep phase (optimizing $\phi$ with respect to the sleep term); the exact schedule for each setting is described below. We use the Adam optimizer and a latent size of 2 in all experiments. On single NVIDIA V100 or A100 GPUs, a run takes about 0.3--2 hours for the baselines, about 2--3 hours for the two stages of unsupervised DDVI, and 7--11 hours for semi-supervised DDVI (batch size 128, one decoder pass per partition). The training details of each algorithm are given below.

\textbf{AEVB.} The batch size is set to 128. The number of epochs is 200 for unsupervised and clustering experiments and 50 for semi-supervised experiments. The learning rate is 0.0001. The loss is BCE for MNIST experiments and MSE for CIFAR-10 and genotype analysis experiments. For semi-supervised experiments, the KL divergence weight is 0.01 for MNIST and 0.5 for CIFAR-10 (the same values as for DDVI in Tables~\ref{tab:semi-supervised} and~\ref{tab:semi-supervised_cifar}), and the classification loss is weighted by the same value. For other experiments, the KL divergence weight increases linearly from 0 to 0.01 over the first 80\% of the epochs. The prior density weight is 5.

\textbf{AEVB-IAF.} The batch size, number of epochs, learning rate, loss, and KL divergence weight are the same as for AEVB; the prior density weight is 1. The context size, i.e., the size of features used to initialize the flow layers for different data points, is 10. In the semi-supervised experiments, the loss of AEVB and AEVB-IAF also contains the term $-0.001\, w_p \beta_\text{reg} H(q_\phi(l|\bx))$ on unlabeled data, where $w_p$ is the prior density weight and $\beta_\text{reg}$ the KL divergence weight; it encourages a higher entropy of $q_\phi(l|\bx)$ (we keep the original implementation; the corresponding term of DDVI below has the opposite sign).

\textbf{H-AEVB.} The batch size, number of epochs, learning rate, and loss are the same as for AEVB; the KL divergence weight increases linearly from 0 to 0.005 (as for AEVB), the prior density weight is 0.5, and the context size is 10.

\textbf{PIS.} The latent follows the controlled SDE $d\bz = u_\phi(\bz, t, \bx)\,dt + \sigma\, d\mathbf{w}$ on $t \in [0, 1]$ from $\bz_0 = \bzero$, with $\sigma = 1$ and 20 Euler--Maruyama steps; without control ($u_\phi = 0$), $\bz_1 \sim \mathcal{N}(\bzero, \sigma^2 \bI)$. By Girsanov's theorem, the ELBO is $\Exp[\log p_\theta(\bx|\bz_1) + \log p(\bz_1) - \log \mathcal{N}(\bz_1; \bzero, \sigma^2 \bI) - \sum_k \lVert u_k \rVert^2 \Delta t / (2\sigma^2)]$, where $u_k$ is the drift at step $k$ and $\Delta t = 1/20$. We optimize it on-policy by backpropagating through the sampled path, and we use the same expression to report the ELBO. The batch size, number of epochs, learning rate, loss, KL divergence weight schedule (applied to all terms but the reconstruction term), and prior density weight (5) are the same as for AEVB.

\textbf{AAEB.} The batch size is set to 128. The number of epochs is 200 for MNIST experiments and 100 for other experiments. The learning rate is 0.0002. The loss is MSE for all experiments. To stabilize the training, we add relative Gaussian noise to the input of the discriminator, with $\sigma = 0.3$ at the start, lowered by 0.1 every quarter of training, so that the noise equals 0 in the last quarter (in the unsupervised experiments, the schedule is shifted by one step, so that the noise equals 0 in the second half of training).

\textbf{DDVI: common settings.} The learning rate is 0.0001. The loss is BCE for MNIST experiments and MSE for CIFAR-10 and genotype analysis experiments. The following settings apply to the simplified algorithm (the first stage of the unsupervised experiments, and the semi-supervised and clustering experiments); the second stage of the unsupervised experiments is described separately below. The forward process $r$ is Gaussian diffusion with a linear schedule $\beta_t = 1 - \alpha_t$ going from $10^{-4}$ to $0.02$, so that $\bar\alpha_T = 0.82$ for $T=20$ and $0.36$ for $T=100$. The noise-prediction loss uses the Huber (smooth-$\ell_1$) loss instead of the squared error. Samples from $q_\phi(\bz|\by_T,\bx)$ are generated with the deterministic DDIM update \citep{song2020denoising}, both during training and evaluation, so that the stochasticity of $q_\phi(\bz|\bx)$ comes from $q_\phi(\by_T|\bx)$. Denoisers that do not depend on $\bx$ (first training stage, semi-supervised, and clustering experiments) are first warmed up on samples from the prior with the noise-prediction loss for $10^5$ iterations (batch size 128, learning rate 0.001).

\textbf{DDVI: unsupervised experiments.} Training proceeds in two stages with batch size 128, for 200 epochs in total. In the first stage (100 epochs, $T=20$, KL divergence weight 0.01), we use the simplified wake-sleep algorithm of Section~\ref{sec:latent-wake-sleep}: the denoiser does not depend on $\bx$, the prior regularization term is approximated by $\KL(q_\phi(\by_T|\bx)||\mathcal{N}(\bzero,\bI))$, the diffusion loss uses samples from $p(\bz)$, and all terms are optimized end-to-end. In the second stage (100 epochs, $T=20$, KL divergence weight 0.003), the encoder trunk and the decoder are initialized from the first stage, and the full wake-sleep algorithm of Section~\ref{sec:latent-wake-sleep} is used with the $\bx$-conditional reverse process of Appendix~\ref{app:model_architecture}. The forward process $r$ (on the standardized latent) has a linear schedule with $\beta_T / \beta_1 = 200$, scaled so that $\bar\alpha_T = \prod_t (1-\beta_t) = 0.5$. The denoiser is first pre-trained on prior samples with a null context ($5 \cdot 10^4$ iterations, batch size 512, learning rate 0.001). In each wake step, we sample a trajectory from $q_\phi$ with the stochastic reverse process (Gaussian steps with the variance of the forward-process posterior, and variance $\beta_1$ for the final step) and maximize a surrogate of $\Lc_\text{rec} + \beta_\text{reg} \Lc_\text{reg}$ along it, with $\beta_\text{reg}$ the KL divergence weight and gradients propagated through the trajectory (as for all methods, the reconstruction term is averaged over pixels during training). In the surrogate, the per-step terms of $\Lc_\text{reg}$ are replaced by the closed-form Gaussian KL divergences between $q_\phi(\by_{t-1}|\by_t,\bx)$ and $r(\by_{t-1}|\by_t,\bz)$ at the sampled $\bz$, the terminal term uses $\mathcal{N}(\bzero,\bI)$ in place of $r(\by_T|\bz)$, and $\log p(\bz)$ is evaluated with the KDE of Appendix~\ref{app:priors}. The decoder is kept fixed during the first 30 epochs of this stage. Each wake step is followed by one sleep step, which updates only the inference networks with a separate Adam optimizer (learning rate 0.0003). We sample $\bz \sim p(\bz)$ and a fantasy $\hat\bx \sim p_\theta(\bx|\bz)$ from the likelihood of the decoder: Bernoulli for MNIST, and Gaussian for CIFAR-10 with its variance set to a running average of the per-pixel squared reconstruction error. We then minimize the squared-error noise-prediction loss of the $\hat\bx$-conditional denoiser (with the context set to zero with probability 0.1) plus $0.05 \cdot \KL(r(\by_T|\bz)||q_\phi(\by_T|\hat\bx))$. The latents used for evaluation are produced by the deterministic DDIM update.

\textbf{DDVI: model selection.} The configuration of the second stage was selected on a validation split: we held out 5,000 training images (a fixed random permutation, the same for all seeds), retrained the first stage on the remaining images, and trained each candidate with seed 0. On MNIST, the candidates combined two recipes (the linear schedule above, or a cosine schedule with $\bar\alpha_T \approx 0$ in which the per-step and terminal terms are weighted separately from the KL divergence weight) with two forms of the context (an encoder head, or a separate network with a 64- or 8-dimensional context). On CIFAR-10, two further rounds varied the fantasy (decoder mean or Gaussian sample), whether the decoder was frozen, the BatchNorm statistics used in the sleep step, the number of sleep steps (1--3), $T \in \{20, 40\}$, $\bar\alpha_T \in \{0.3, 0.5\}$, an additional trajectory-matching term, the size of the denoiser, and the form of the context. We selected the candidate with the best rank, averaged over the metrics reported for the dataset (ELBO, MMD, and Latent NLL for MNIST; ELBO and Latent NLL for CIFAR-10) and over the three priors, on the validation split; this rule was fixed before the search. The same configuration, described above, was selected on both datasets. Only DDVI was tuned in this way; the baselines use the settings of this appendix. The candidates were designed in earlier development runs that were evaluated on the test set, and on CIFAR-10 the two further rounds followed a test-set evaluation of the first-round choice; the test set was not used to choose among the candidates. On CIFAR-10, the selection was carried out with the earlier sigmoid decoder (Appendix~\ref{app:model_architecture}) and was not repeated with the tanh decoder used for all methods in Tables~\ref{tab:unsupervised_cifar} and~\ref{tab:semi-supervised_cifar}.

\textbf{DDVI: semi-supervised experiments.} The batch size is 128, $T=100$, and the number of epochs is 50, with one denoiser per partition $p(\bz|l)$ of the prior. These denoisers do not depend on $\bx$: they are trained with the simplified sleep term (the noise-prediction loss on samples from their partitions) during the warm-up and then kept fixed. The encoder, the label posterior $q_\phi(l|\bx)$, and the decoder are trained with the reconstruction term averaged over pixels and over the batch, as for AEVB and AEVB-IAF; the prior regularization term, approximated by $\KL(q_\phi(\by_T|\bx)||\mathcal{N}(\bzero,\bI))$ and multiplied by the KL divergence weight; and the classification loss of Section~\ref{sec:semi}, with $\alpha$ equal to the KL divergence weight. As for the baselines, the decoder does not depend on $l$, i.e., $p_\theta(\bx|\bz,l) = p_\theta(\bx|\bz)$. The KL divergence weight is 0.01 for MNIST and 0.5 for CIFAR-10, the same as for AEVB and AEVB-IAF in each table (0.01 is the baselines' original value, and 0.5 is DDVI's original value for CIFAR-10). In an earlier version of this paper, the reconstruction term of DDVI was normalized differently (labeled and unlabeled data were averaged separately, and the sum was divided by the sum rather than the product of the image dimensions), which reduced its effective KL divergence and classification weights by a factor of about 2 on MNIST and about 93 on CIFAR-10. We also add to the loss an entropy-minimization term $0.001 \cdot H(q_\phi(l|\bx))$ on unlabeled data, multiplied by the KL divergence weight. The latents of all partitions are decoded in a single batch, so that the BatchNorm statistics of the CIFAR-10 decoder are computed over all classes; decoding each partition separately (as in an earlier version of this paper) gives class-homogeneous batches whose statistics differ from those used at evaluation.

\textbf{DDVI: clustering experiments.} The batch size is 128, the number of epochs is 200, $T=20$, and the KL divergence weight is 0.005. We use the simplified wake-sleep algorithm with the mixture prior of Section~\ref{sec:clustering} (20 components with learnable means). After each epoch $e$, we take $e+1$ gradient steps on the noise-prediction loss of the denoiser (learning rate 0.001) and $e+1$ gradient steps on the cluster means (learning rate 0.01). The means are trained with the noise-prediction loss on prior samples, a hinge penalty that keeps each pair of means at least 1.5 apart (with a weight that increases after epoch 100), and a penalty that keeps them within $[-2.5, 2.5]^2$.

\textbf{Evaluation.} We report the ELBO $\Exp_{q_\phi(\bz|\bx)}[\log p_\theta(\bx|\bz) + \log p(\bz) - \log q_\phi(\bz|\bx)]$ in nats per test image, estimated with one sample of $\bz$ per image over the test set. The reconstruction term is summed over pixels: for MNIST, the Bernoulli log-likelihood (binary cross-entropy) of the grey-level pixels, which, as is common, is not normalized for non-binary data; for CIFAR-10, a Gaussian likelihood $\mathcal{N}(\bx; \mu_\theta(\bz), \hat\sigma^2 \bI)$, where $\hat\sigma^2$ is fit by maximum likelihood for each trained model (the mean per-pixel squared reconstruction error on the training set). For all methods, $\log p(\bz)$ is evaluated with the KDE of Appendix~\ref{app:priors}. The term $\log q_\phi(\bz|\bx)$ is exact for AEVB and AEVB-IAF; H-AEVB uses the auxiliary-variable bound of Appendix~\ref{app:elbo_aux} with a standard normal reverse model, and PIS uses the bound above. These two bounds involve auxiliary variables and are looser than the ELBO of the corresponding marginal posteriors. For DDVI, the latents are produced by the deterministic DDIM update, so $q_\phi(\bz|\bx)$ is the pushforward of $q_\phi(\by_T|\bx)$ by a map $F_\bx$, and $\log q_\phi(\bz|\bx) = \log q_\phi(\by_T|\bx) - \log|\det \partial F_\bx(\by_T)/\partial \by_T|$, where the $2\times2$ Jacobian is computed exactly by automatic differentiation ($F_\bx$ includes the standardization of the latent). This is exact provided that $F_\bx$ is injective, which we assume but have not verified for the learned finite-step map. Where dropout is active at evaluation, the dropout masks are treated as auxiliary variables. In the semi-supervised setting, the ELBO of all methods also includes $-\KL(q_\phi(l|\bx)||p(l))$ with a uniform $p(l)$. In an earlier version of this paper, the reconstruction term was averaged over pixels and the ELBO of DDVI was computed with an estimator that is not a lower bound (it omitted the entropy of the last reverse step and treated the deterministic DDIM steps as Gaussian transitions with the variance of the forward-process posterior); all ELBOs are now computed as described here. Acc denotes the test accuracy of a $20$-nearest-neighbor classifier fit on the latents of all training images and their labels.

\section{Genotype Analysis Experiment Details}\label{app:genotype}

\begin{figure*}[t]
    \centering
    \includegraphics[width=1\textwidth]{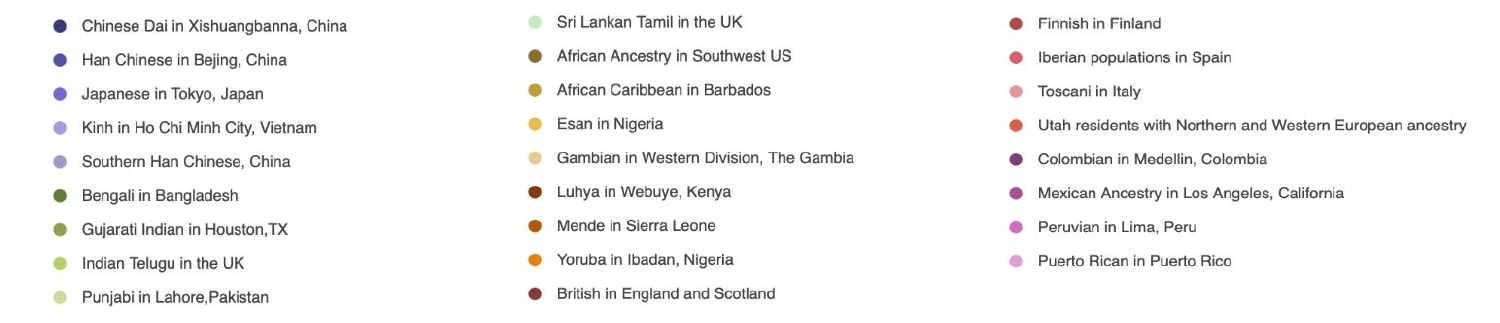}
    \vspace{-1em}
    \caption{Populations of the 1000 Genomes dataset and their colors in Figure~\ref{fig:genotype}.}
    \vspace{-10pt}
    \label{fig:legend}
\end{figure*} 
Before inputting the data points into any of the visualization methods, we first pre-process them with PCA, keeping only the first 1000 principal components of the data points. We further divide the features by 30 for all latent variable models. For all latent variable models, cluster assignments are obtained by fitting a 20-component Gaussian mixture model (full covariance, 10 initializations) to the 2-D latents of all individuals, and cluster purity, completeness, and NMI are computed with respect to the population labels. 

The legend of the 1000 Genomes visualization can be found in Figure \ref{fig:legend}.

\begin{table*}[ht]
\centering
\adjustbox{max width=\textwidth}{
\centering
\setlength\tabcolsep{4pt} %
\begin{tabular}{c@{\hspace{5pt}}c@{\hspace{5pt}}c@{\hspace{5pt}}c@{\hspace{5pt}}c@{\hspace{5pt}}c@{\hspace{5pt}}c}
\toprule
Method & \multicolumn{2}{c}{Pinwheel} & \multicolumn{2}{c}{Swiss Roll} & \multicolumn{2}{c}{Square}\\ 
\cmidrule(lr){2-3} \cmidrule(lr){4-5} \cmidrule(lr){6-7}  
& ELBO & Latent NLL & ELBO & Latent NLL & ELBO & Latent NLL \\ 
\midrule
AEVB & $-1608.1 \pm 40.9$ & $2.32 \pm 0.17$ & $-1522.1 \pm 68.2$ & $6.79 \pm 0.66$ & $-1578.3 \pm 23.0$ & $2.53 \pm 2.07$\\ 
AEVB-IAF & $-1491.2 \pm 29.3$ & $1.81 \pm 1.13$ & $-1464.7 \pm 35.6$ & $5.24 \pm 1.54$ & $-1475.8 \pm 37.9$ & $1.54 \pm 0.23$\\ 
AAEB & $N/A$ & $1.40 \pm 0.21$ & $N/A$ & $2.86 \pm 0.12$ & $N/A$ & $1.80 \pm 0.32$\\ 
H-AEVB & $\mathbf{-1401.8 \pm 13.9}$ & $1.46 \pm 0.02$ & $\mathbf{-1401.4 \pm 20.7}$ & $5.97 \pm 0.14$ & $-1403.6 \pm 24.6$ & $1.60 \pm 0.05$\\ 
PIS$^\dagger$ & $-1686.6 \pm 15.7$ & $1.88 \pm 0.02$ & $-1991.6 \pm 179.6$ & $6.01 \pm 0.23$ & $-1865.2 \pm 101.9$ & $2.72 \pm 1.59$\\ 
DDVI & $-1453.2 \pm 24.9$ & $\mathbf{0.97 \pm 0.07}$ & $-1548.9 \pm 58.7$ & $\mathbf{2.46 \pm 0.31}$ & $\mathbf{-1363.1 \pm 27.2}$ & $\mathbf{1.50 \pm 0.06}$\\ 
\bottomrule
\end{tabular}}
\caption{Unsupervised learning on CIFAR-10 (3 seeds): ELBO (nats per image; Gaussian decoder with maximum-likelihood variance, Appendix~\ref{app:training_details}) and Latent NLL. The configuration of DDVI was selected on a validation split with the earlier sigmoid decoder (Appendix~\ref{app:training_details}). $^\dagger$Our reimplementation (Appendix~\ref{app:model_architecture}).}
\label{tab:unsupervised_cifar}
\end{table*}
\begin{table*}[ht]
\centering
\adjustbox{max width=\textwidth}{
\centering
\setlength\tabcolsep{4pt} %
\begin{tabular}{c@{\hspace{5pt}}c@{\hspace{5pt}}c@{\hspace{5pt}}c@{\hspace{5pt}}c@{\hspace{5pt}}c@{\hspace{5pt}}c@{\hspace{5pt}}c@{\hspace{5pt}}c@{\hspace{5pt}}c}
\toprule
Method & \multicolumn{3}{c}{Pinwheel} & \multicolumn{3}{c}{Swiss Roll} & \multicolumn{3}{c}{Square}\\ 
\cmidrule(lr){2-4} \cmidrule(lr){5-7} \cmidrule(lr){8-10}  
& ELBO & Acc & Latent NLL & ELBO & Acc & Latent NLL & ELBO & Acc & Latent NLL \\ 
\midrule
AEVB & $-2113.7 \pm 44.4$ & $0.30 \pm 0.07$ & $2.24 \pm 0.18$ & $-2167.4 \pm 82.0$ & $0.20 \pm 0.07$ & $6.52 \pm 2.12$ & $-2125.1 \pm 19.1$ & $0.30 \pm 0.05$ & $1.94 \pm 0.28$\\ 
AEVB-IAF & $-2090.1 \pm 29.9$ & $0.48 \pm 0.02$ & $1.60 \pm 0.08$ & $-2093.7 \pm 55.8$ & $0.29 \pm 0.03$ & $7.09 \pm 0.31$ & $-2118.9 \pm 21.3$ & $0.35 \pm 0.01$ & $1.57 \pm 0.04$\\ 
AAEB$^\ddagger$ & $N/A$ & $0.26 \pm 0.01$ & $1.70 \pm 0.03$ & $N/A$ & $0.25 \pm 0.03$ & $3.17 \pm 0.19$ & $N/A$ & $0.24 \pm 0.02$ & $1.67 \pm 0.28$\\ 
DDVI & $\mathbf{-2053.1 \pm 14.9}$ & $\mathbf{0.53 \pm 0.02}$ & $\mathbf{1.07 \pm 0.01}$ & $\mathbf{-2069.3 \pm 15.6}$ & $\mathbf{0.52 \pm 0.02}$ & $\mathbf{2.12 \pm 0.03}$ & $\mathbf{-2065.4 \pm 19.3}$ & $\mathbf{0.50 \pm 0.01}$ & $\mathbf{1.50 \pm 0.02}$\\ 
\bottomrule
\end{tabular}}
\caption{Semi-supervised learning on CIFAR-10 (10,000 labels; 3 seeds): ELBO (nats per image; Appendix~\ref{app:training_details}), 20-NN accuracy (Acc), and Latent NLL. All methods use KL and classification weights 0.5, batch size 128, and 50 epochs ($^\ddagger$except AAEB; Appendix~\ref{app:training_details}).}
\label{tab:semi-supervised_cifar}
\end{table*}
 
\begin{table*}[ht]
\centering
\adjustbox{max width=\textwidth}{
\centering
\setlength\tabcolsep{4pt} %
\begin{tabular}{c@{\hspace{5pt}}c@{\hspace{5pt}}c@{\hspace{5pt}}c}
\toprule
Method & Latent NLL - Pinwheel & Latent NLL - Swiss Roll & Latent NLL - Square\\
\midrule
AEVB & $1.55 \pm 0.59$ & $5.63 \pm 1.72$ & $2.93 \pm 1.11$\\ 
AEVB-IAF & $1.29 \pm 0.40$ & $3.70 \pm 1.97$ & $1.72 \pm 0.10$\\ 
AAEB & $1.54 \pm 0.13$ & $3.34 \pm 0.24$ & $2.27 \pm 0.87$\\ 
H-AEVB & $1.54 \pm 0.11$ & $2.92 \pm 0.45$ & $1.67 \pm 0.04$\\ 
DDVI & $\mathbf{1.11 \pm 0.10}$ & $\mathbf{2.27 \pm 0.16}$ & $\mathbf{1.56 \pm 0.01}$\\ 
\textit{DDVI (w/o sleep step)} & $3.29 \pm 0.92$ & $6.99 \pm 4.67$ & $2.39 \pm 0.62$\\ 
\bottomrule
\end{tabular}}
\caption{Unsupervised learning on MNIST (Latent NLL; 3 seeds), including DDVI trained without the sleep step of the second training stage (all other settings as in Table~\ref{tab:unsupervised}).}
\vspace{-10pt}
\label{tab:unsupervised_w_failed}
\end{table*}
 
\section{ELBO for Auxiliary-Variable Generative Models}\label{app:elbo_aux}

We aim to derive the lower bound for the log-likelihood $\log p_\theta(\bx)$ by introducing auxiliary variables and applying the Evidence Lower Bound (ELBO) twice.

Introduce the latent variable $\bz$ and apply the ELBO:
\begin{align}
    \log p_\theta(\bx) 
    &= \log \int p_\theta(\bx, \bz) \, d\bz \\
    &\geq \mathbb{E}_{q_\phi(\bz|\bx)} \left[ \log \frac{p_\theta(\bx, \bz)}{q_\phi(\bz|\bx)} \right] \quad \text{(by Jensen's inequality)} \\
    &= \mathbb{E}_{q_\phi(\bz|\bx)} \left[ \log p_\theta(\bx|\bz) \right] - \text{KL}\left( q_\phi(\bz|\bx) \,\|\, p(\bz) \right)
\end{align}

Then, introduce an auxiliary variable $\by$ and apply the ELBO again:
\begin{align}
    \log p_\theta(\bx) 
    &\geq \mathbb{E}_{q_\phi(\bz|\bx)} \left[ \log p_\theta(\bx|\bz) \right] - \text{KL}\left( q_\phi(\bz|\bx) \,\|\, p(\bz) \right) \\
    &= \mathbb{E}_{q_\phi(\bz|\bx)} \left[ \log p_\theta(\bx|\bz) \right] - \mathbb{E}_{q_\phi(\bz|\bx)} \log q_\phi(\bz|\bx) + \mathbb{E}_{q_\phi(\bz|\bx)} \log p(\bz) \\
    &\geq \mathbb{E}_{q_\phi(\bz|\bx)} \left[ \mathbb{E}_{q_\phi(\by|\bx,\bz)} \left[ \log p_\theta(\bx|\bz) \right] \right] - \mathbb{E}_{q_\phi(\bz|\bx)} \left[ \mathbb{E}_{q_\phi(\by|\bx, \bz)} \left[ \log \frac{q_\phi(\by, \bz|\bx)}{r(\by|\bx, \bz)} \right] \right] + \mathbb{E}_{q_\phi(\bz|\bx)} \left[ \mathbb{E}_{q_\phi(\by|\bx,\bz)} \left[ \log p(\bz) \right] \right] \\
    &= \mathbb{E}_{q_\phi(\by,\bz|\bx)} \left[ \log p_\theta(\bx|\bz) \right] - \text{KL}\left( q_\phi(\by,\bz|\bx) \,\|\, r(\by|\bx,\bz) p(\bz) \right) 
\end{align}

\section{Diffusion Regularization}\label{app:diff_reg}

We begin with the definition of \(\Lc_\text{diff}(\phi)\), derived as a lower bound on \(\Lc_\text{sleep}(\phi)\):
\[
\Lc_\text{diff}(\phi) = \Exp_{p_\theta(\bx,\bz)} \left[ \Exp_r \left[ \log \frac{q_\phi(\by, \bz | \bx)}{r(\by | \bz, \bx)} \right] \right] + \bar{H}(p_\theta).
\]

To simplify \(\Lc_\text{diff}\), we leverage the Markov structure of the forward process \(r(\by | \bz, \bx)\) and the reverse process \(q_\phi(\by, \bz | \bx)\).

The forward process \(r(\by | \bz, \bx)\) is decomposed as:
\[
r(\by_{1:T} | \bz, \bx) = \prod_{t=1}^T r(\by_t | \by_{t-1}, \bx),
\]
where \(\by_0 = \bz\).

The reverse process \(q_\phi(\by, \bz | \bx)\) is decomposed as:
\[
q_\phi(\by, \bz | \bx) = q_\phi(\by_{0:T} | \bx) = q_\phi(\by_T | \bx) \prod_{t=1}^T q_\phi(\by_{t-1} | \by_t, \bx),
\]
where \(\by_0 = \bz\) and \(\by_{1:T}\) are increasingly noisy versions of \(\bz\).

Substituting the factorizations of \(r\) and \(q_\phi\) into the definition of \(\Lc_\text{diff}(\phi)\), and rewriting the logarithm and rearranging terms, we have:
\begin{align*}
\Lc_\text{diff}(\phi) &= \Exp_{p_\theta(\bx, \bz)} \Bigg[ \Exp_{r(\by_{1:T} | \bz, \bx)} \Bigg[ \log \frac{q_\phi(\by_T | \bx) \prod_{t=1}^T q_\phi(\by_{t-1} | \by_t, \bx)}{\prod_{t=1}^T r(\by_t | \by_{t-1}, \bx)} \Bigg] \Bigg] + \bar{H}(p_\theta) \\
& = \Exp_{p_\theta(\bx, \bz) r(\by_{1:T} | \bz, \bx)} \left[\log q_\phi(\bz | \by_1, \bx) - \sum_{t=2}^T \KL(r_t || q_t) - \KL(r(\by_T | \bz, \bx) || q_\phi(\by_T | \bx))\right] + \bar{H}(p_\theta)
\end{align*}

Here, \(\log q_\phi(\bz | \by_1, \bx)\) corresponds to the reconstruction term for the initial latent state \(\bz\), while the summation represents the KL divergence between the forward and reverse processes at each intermediate step. \(r_t = r(\by_{t-1} | \by_t, \bz, \bx)\) is the posterior of the forward process at step \(t\), and \(q_t = q_\phi(\by_{t-1} | \by_t, \bx)\) is the reverse process distribution at step \(t\). $\bar H(p_\theta)$ is the expected conditional entropy of $p_\theta(\bz|\bx)$, a constant that does not depend on $\phi$.

This form mirrors the ELBO derivation for diffusion models \citep{sohl2015deep}, where each step in the Markov chain contributes a KL divergence term, and the reconstruction term arises from the connection between the noisy latent variables and the original data.

\end{document}